\documentclass[twocolumn,preprintnumbers,superscriptaddress]{revtex4-2}
\usepackage{hyperref}
\usepackage[utf8]{inputenc}

\usepackage{soul}

\usepackage{url}

\usepackage{natbib}
\usepackage{graphicx}
\usepackage{physics}
\usepackage{tikz}
\usetikzlibrary{quantikz}

\newcommand{\DSD}{
Research Group on Data Science for the Digital Society
La Salle - Universitat Ramon Llull
Carrer de Sant Joan de La Salle, 42
08022 Barcelona (Spain)
}
\newcommand{\LDIG}{
Lighthouse Disruptive Innovation Group, LLC
7 Broadway Terrace, Apt 1
Cambridge MA 02139
Middlesex County, Massachusetts (USA)
}
\newcommand{\orcid}[1]{\href{https://orcid.org/#1}{\includegraphics[width=8pt]{orcid.png}}}

\renewcommand{\arraystretch}{2} 

\graphicspath{{./images/}}

\begin{document}

\title{quantum Case-Based Reasoning (qCBR)}
\author{Parfait Atchade Adelomou}
\affiliation{\DSD}
\email{parfait.atchade@salle.url.edu}
\affiliation{\LDIG}

\author{Daniel Casado Fauli}
\affiliation{\DSD}
\email{daniel.casado@salle.url.edu}
\author{Elisabet Golobardes Ribé}
\affiliation{\DSD}
\email{elisabet.golobardes@salle.url.edu}
\author{Xavier Vilas\'\i s-Cardona}
\affiliation{\DSD}
\email{xavier.vilasis@salle.url.edu}

\date{March 2021}

\begin{abstract}
Case-Based Reasoning (CBR) is an artificial intelligence approach to problem-solving with a good record of success. This article proposes using Quantum Computing to improve some of the key processes of CBR, such that a Quantum Case-Based Reasoning (qCBR) paradigm can be defined.
The focus is set on designing and implementing a qCBR based on the variational principle that improves its classical counterpart in terms of average accuracy, scalability and tolerance to overlapping. A comparative study of the proposed qCBR with a classic CBR is performed for the case of the Social Workers' Problem as a sample of a combinatorial optimization problem with overlapping. The algorithm's quantum feasibility is modelled with docplex and tested on IBMQ computers, and experimented on the Qibo framework.
\end{abstract}

\maketitle

\section{Introduction}\label{sec:introduction}
The social workers' problem (SWP) stands for solving the schedules of social workers visiting patients' homes while fitting both distance and time restrictions~\cite{Atc20} and represents a class of combinatorial optimization problems, which lie in the NP complexity class.
The standard way to solve this class of problems begins by establishing the cost function. Then, depending on its form, existing linear or quadratic programming methods such as Simplex \cite{Simplex_solver} or Cplex \cite{CPlex_solver} can be applied. More complex cost functions require more sophisticated numerical methods. Depending on the problem's complexity class, the algorithm can be improved by introducing some heuristics or restrictions in the objective function to reduce its computational cost for an approximate solution. When the size of the problem grows, the computational cost may soon become intractable for the current computational paradigms. In addition to the above, solving these problems is more challenging when the input data presents some overlapping issues or when outstanding accuracy is required. 

In this paper, an approach combining adapting Case-Based Reasoning\cite{Aam94} to a quantum computing is proposed to solve this class of problems. This paradigm, denoted Quantum Case-Based Reasoning (qCBR), will address both the overlap in the input data and the accuracy problem. Furthermore, by directly constructing the framework, questions like the actual efficiency of a qCBR implementation at the present level of quantum technology, the tolerance concerning input overlap, the scalability and the applicability to other combinatorial optimization problems will be discussed. 
The paper is organized as follows. Section \ref{sec:Related_work}, shows previous work on both ensemble techniques and quantum machine learning approaches. In Section \ref{sec:Case_based_reasoning}, the Case-Based Reasoning will be explored by focusing on its features. Section \ref{sec:QCircuit_NISQ} introduces the necessary quantum fundamentals for this era to solve this problem. Implementing the proposed strategy and creating the quantum CBR (qCBR) is done in Section \ref{sec:implementation}. In Section \ref{sec:result}, we present our experimental analysis results. In Section \ref{sec:Discussions}, we summarise, benchmark and present some open problems. Finally, Section \ref{sec:Conclusions} concludes previous results and outlines future work.

\section{Related Work}\label{sec:Related_work}
CBR is a problem solving approach widely considered in the literature with a large record of success. Application examples are a medical reasoning program that improves with experience \cite{koton1989medical}, an individual prognosis of diabetes long-term risks\cite{armengol2001individual}, Case-Based Sequential ordering of songs for playlist recommendation \cite{baccigalupo2006case}, ranking order in financial distress prediction \cite{li2008ranking}, monitoring the elderly at home \cite{koton1989medical}, software control \cite{Aam94}, in the medical field\cite{Aam94}, sequencing problems \cite{Pao01} etc.
In one of our previous works \cite{Atc201,Atc20}, we observed part of the benefits of using the CBR instead of the Top-Down method. And the needs of empowering this problem-solving method based on human learning were seen.

Quantum computing stands as a new computing paradigm based on exploiting the principles of quantum mechanics and establishing the quantum bit (qubit) as the elementary unit of information. It emerged in the early 1990's from algorithms that were able to take advantage of quantum characteristics to show advantages over their classical counterparts, being Shor's algorithm \cite{shor1994algorithms} for integer factorization and Grover's algorithm \cite{grover1996fast} for searching in an unordered data sequence, the most famous. However, current quantum computing devices suffer from technological limitations, such as the number of qubits available and the noise and decoherence problems, such that they are still no match for their classical counterparts. This situation is known as the Noisy Intermediate-Scale Quantum (NISQ) era \cite{Joh18}. These limitations have forced the scientific community to develop handy tools for hybrid computing, mixing classical and quantum. Taking advantage of the variational principle, it is possible to solve combinatorial optimization problems and enhance one of this era's most promising fields; quantum machine learning (QML) \cite{Mar14,Adr20}. In this new approach, several techniques and methods already explored in Machine Learning (ML) are being worked on.

In the last two years, the number of algorithms based on QML have increased considerably since the first definition in 2014 \cite{Mar14}. This progress relies on the advances in decoherence  control \cite{decoherence_ctrl,Quantum_decoherence} and error correction systems \cite{Quantum_error_correct} combined with the availability of several quantum server providers in the cloud. Most of these new algorithms take after the variational principle, being the Variational Quantum EigenSolver (VQE) \cite{Peruzzo2014} and the Quantum Approximate Optimization Algorithm (QAOA) \cite{farhi2014quantum,QAOA_2019,Adapt_QAOA} the most famous. Other promising developments are the Quantum Neural Network (QNN) \cite{QNN_2014,Quest_QNN_2014,Train_QNN_2020}, the Quantum Support Vector Machine (QSVM)\cite{Pat14,SL_Quantum_Feature_Space,SVM_DWAVE} and the data loading system \cite{UAT2021,Mar19}. On the one hand, the following references \cite{a14070194, gonzalezbermejo2021gps} highlight works done in the Top-Down philosophy. On the other hand, references \cite{zhao2021smooth, lamata2020quantum, benfenati2021improved, cerezo2021cost, alonsolinaje2021eva, atchadeadelomou2021quantum} highlight the many contributions in quantum machine learning, from using the properties of quantum computing to finding new drugs as new ways to calculate the expected value, among others.

The literature shows examples of exploiting the possibilities of hybrid (classical-quantum) computing connecting it to CBR. For instance, in reference~\cite{Cognite_CBR_GA} a cognitive engine that uses CBR-QGA to adjust and optimize the radio parameters is presented. An initial quantum bit made up of the matching case parameters is used to avoid blindness of the initial population search and speed up optimization of the quantum genetic algorithm. References~\cite{Modeling_CBR,CBR_Ontology} propose a new framework that can be adopted in many applications that require Computational Intelligence (CI) solutions. The framework is built under the concepts of Soft Computing (SC), where Fuzzy Logic (FL), Artificial Neural Network (ANN) and Genetic Algorithm (GA) are exploited to perform reasoning tasks based on soft cases. Also studies \cite{review_KNN_CL_QC} focused on some vital blocks of the CBR were reviewed. It has focused on the quantum version of the k-NN algorithm that allows us to understand the fundamentals when transcribing classic machine learning algorithms into their quantum versions. 

Reviewing state of the art, we have seen an interesting field known as Quantum Information Retrieval (QIR) \cite{piwowarski2010can, lebedev2020introductory, kitto2012quantum} that uses the Gleason theorem \cite{GleasonTheorem} on the Measures on the Closed Subspaces of a Hilbert Space for information retrieval geometry \cite{van2004geometry}. It calculates the probability algebraically through the density matrix trace and acts on a quantum projector. The projector can be any concept to recover. However, for the quantum CBR, we are not only interested in a great recovery system, but we also need to provide the qCBR with a synthesiser whose function will be to fine-tune the recovered data in the case of not being the optimal result since the qCBR has the process of "generate" a new outcome based on the retrieved information.

However, no quantum Case-Based Reasoning was found that can satisfy the requirements presented above. Such is the purpose of this paper. 

\begin{figure}[b!]
\centering
\includegraphics[width=0.5\textwidth]{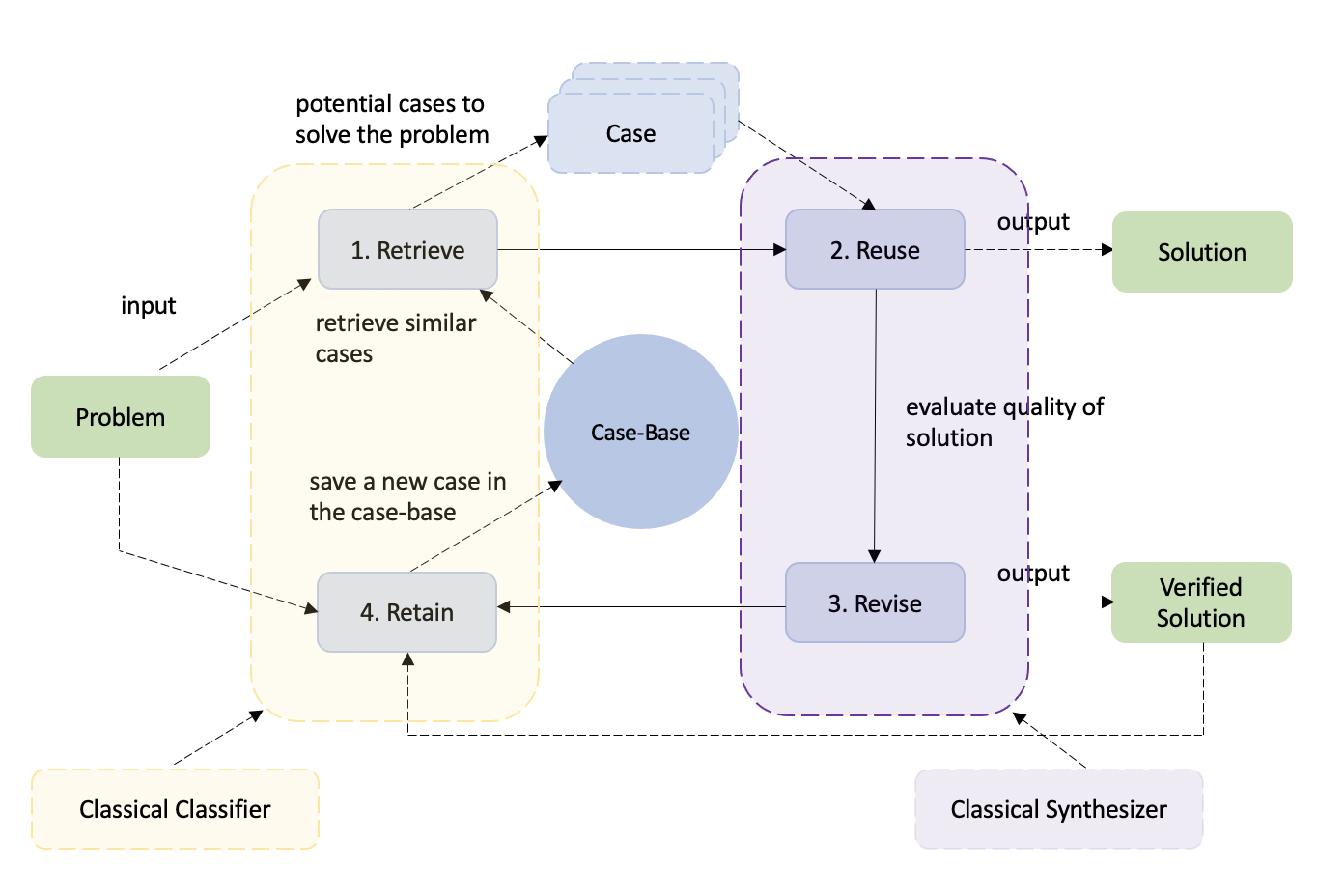}
\caption{Case-Based Reasoning block diagram. In this work, for the standard CBR, two essential blocks are distinguished: The classifier and the synthesizer. The classifier is made up of the retrieve and retain blocks and the re-use and revise blocks to make up the synthesis system.}
\label{fig:CBR}
\end{figure}

\begin{table*}[!t]
    \centering
    \begin{tabular}{c|c|c|c|c}
        \textbf{Methods} & \textbf{ Brute force } & \textbf{ k-d tree method\cite{Kak05} } & \textbf{ Ball tree method\cite{Mun11} } \\
         \hline
        Training time complexity & $O(1)$ & $O(dNlog(N))$ & $O(dNlog(N))$ \\
        Training space complexity & $O(1)$ & $O(dN)$ & $O(dN)$ \\
        Prediction time complexity & $O(KNd)$  & $O(Klog(N))$  & $O(Klog(N))$ \\
        Prediction space complexity & $O(1)$ & $O(1)$ & $O(1)$ \\
    \end{tabular}
    \caption{Table of the NN brute force's, k-d tree's, and Ball tree's complexity method. Where $d$, is the data dimensionality, $N$ is the number of points in the training dataset and $K$ is the algorithm's neighbours' number.}
    \label{tab:Complexity_Classifier}
\end{table*}

\section{Case-Based Reasoning}\label{sec:Case_based_reasoning}
CBR \cite{Aam94} is a machine learning technique based on solving new problems using experience, as humans do. The \textit{experience} is represented as a case memory containing previously solved cases. The CBR cycle can be summarised in four steps: (1) Retrieval of the most similar cases, (2) Adaptation to those cases to propose a new solution to the new environment, (3) Validity check of the proposed solution and finally, (4) Storage following a learning policy. In the present work, the proposed qCBR modifies these phases as follows (see Fig.\eqref{fig:CBR} and \eqref{fig:qCBR}).

The CBR technique could be summarized in two large blocks according to their functionality: a classifier and a synthesizer. One of the classical CBR advantages is its classifier's simplicity, being a k-nearest neighbors algorithm (K-NN)\cite{JHF75,Fuk75} classifier a common option. This apparent advantage can lead to collateral problems \cite{Lau11} at the memory level, at the level of slowness when the volume of data grows considerably and at data synthesis. The synthesis block is in charge of adapting the experience and saving the new problem. Such adaptation and classification can be costly (Table \ref{tab:Complexity_Classifier}) for considerably high data volumes \cite{Abd15}. From this follows that a different approach would be required to further empowering this technique.

The proposal of this note is to achieve such empowering in two steps. First by making a CBR with a quantum classifier \cite{Suk191} instead of a classical neural network, KNN\cite{JHF75,Fuk75} or a Support Vector Machine (SVM) \cite{Wil06} since quantum classifiers offer outstanding accuracy and tolerate overlapping problems \cite{8715261}. The second would be changing the classical synthesis technique for the Variational Quantum Eigensolver (VQE) \cite{Alb13,Dao19,Placeholder1} with \textit{Initial\_point}\cite{Qis21}.

\section{Quantum Circuits in the NISQ era}\label{sec:QCircuit_NISQ}
Quantum circuits are mathematically defined as operations on an initial quantum state. Quantum computing generally makes use of quantum states built from qubits, that is, binary states represented as  $\ket{\psi}=\alpha\ket{0} +\beta\ket{1}$. Their number of qubits $n$  commonly defines the states of a quantum circuit and, in general, the circuit's initial state $\ket\psi_{0}$ is the zero state $\ket{0}$. In general, a quantum circuit implements an internal unit operation $U$  to the initial state $\ket\psi_{0}$ to transform it into the final output state $\ket\psi_{f}$ . This gate $U$  is wholly fixed and known for some algorithms or problems. In contrast, others define its internal functioning through a fixed structure, called Ansatz\cite{Ansatz_best} (Parametrized Quantum Circuit (PQC)), and adjustable parameters $\theta$ \cite{Suk191}. Parameterized circuits are beneficial and have interesting properties in this quantum age since they broadly define the definition of ML and provide flexibility and feasibility of unit operations with arbitrary precision \cite{JBi17,Adr20,Mar14}.

\begin{figure}[!ht]
\centering
\includegraphics[width=0.45\textwidth]{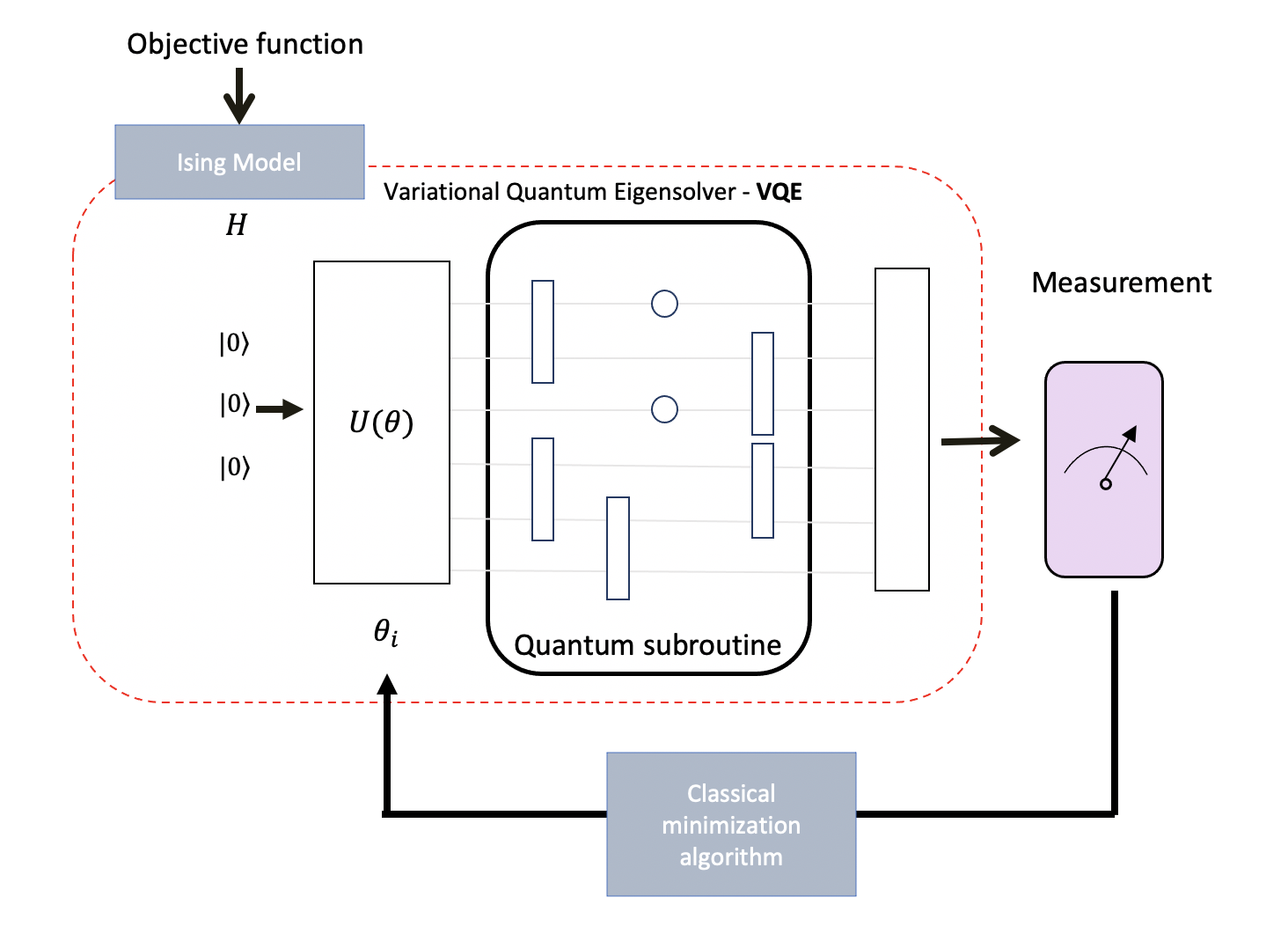}
\caption{VQE working principle based on the quantum variational circuit. Given an objective function that characterises the problem, with the help of the Ising Model block, we pass the objective function from the classical to the quantum domain.
The ansatz is initialised with random values. Then, starting from these initial values (initial position) and depending on the measured value, a classical and external optimiser is used to feedback the new values of the ansatz parameters. So, until reaching the minimum energy value, equivalent to the ground state of our Hamiltonian, defined by the variational principle.}
\label{fig:VQC}
\end{figure}

Figure \eqref{fig:VQC} depicts the concept of hybrid computing (quantum + classical), which defines the NISQ. This takes advantage of quantum computing's capacity to solve complex problems, and the experience of classical optimization algorithms (COBYLA \cite{The21}, SPSA \cite{Jam01}, BFGS \cite{BFGS_Limted}, etc.) to train variational circuits. Classical algorithms are generally an iterative scheme that searches for better candidates for the parameters $\theta$ at each step.

The value of the hybrid computing idea in the NISQ era is necessary because it allows the scientific community to exploit the powers of both and reap the benefits of the constant acceleration of the oncoming quantum-computer development. With a good optimization system and a closed-loop system, the non-systematic noises could be automatically corrected during the optimization process.

Furthermore, with the insertion of information (data) into the variational circuit through the quantum gate $U$, learning techniques can be improved.

The Variational Quantum Circuit (VQC) \cite{Mic00,Mic}, consists of a quantum circuit that defines the base structure similar to neural network architecture (Ansatz) while, the variational procedure can optimize the types of gates (one or two-qubit parametric gates) and their free parameters.
All this is summarized in a few very identifiable steps. 
First, the Ansatz must be designed, using a set of one- and two-qubit parametric gates. The Ansatz of this circuit can follow a particular path by exploiting the problem's characteristics. A critical block is measuring the quantum state resulting from the given Ansatz. Since the VQC is a feedback system, these measurements evaluate a cost function  $CF(\theta)$(see in the Appendix \ref{sec:App-Ansatz}) that encodes the problem. The classical optimizer has the role of optimizing the cost function to find the value of the parameters that minimize it.

\section{Implementation}\label{sec:implementation}
\begin{figure}[!ht]
\centering
\includegraphics[width=0.5\textwidth]{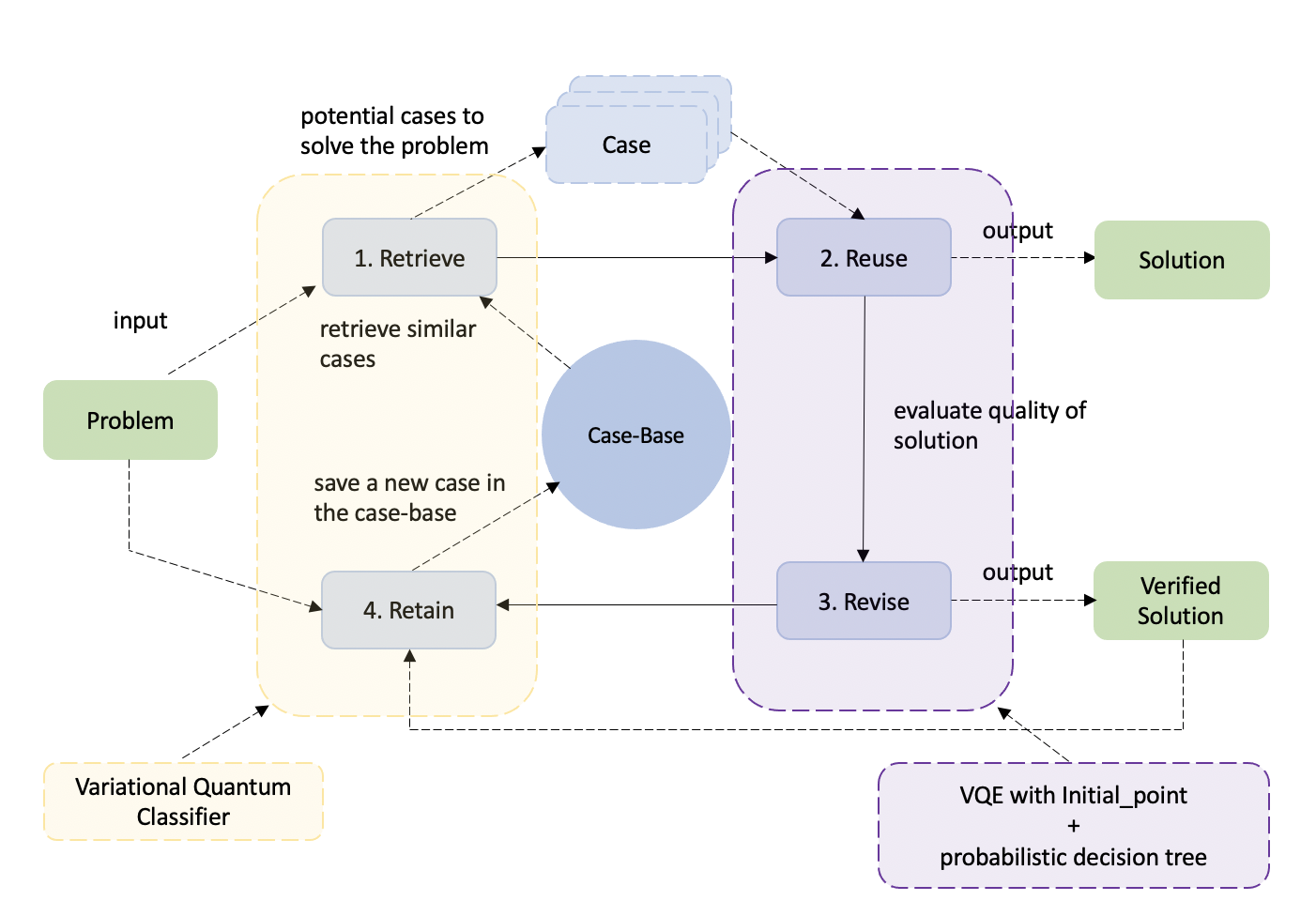}
\caption{Quantum Case-Based system block diagram. In this scheme, to convert the classic CBR into the quantum one, it is proposed to change the retrieve and retain blocks for a quantum variational classifier and the re-use and revise blocks for a synthesis system based on VQE with initial\_point.}
\label{fig:qCBR}
\end{figure}

The work proposed in this article is the implementation of a quantum Case-Based Reasoning (qCBR) based on figure \eqref{fig:CBR}. The strategy to follow is to replace the classical classification: an Artificial Neural Network (ANN) or a Support Vector Machine (SVM) or the KNN with a quantum variational classifier that guarantees the required accuracy. And for the quantum synthesis system, use the VQE with and without \textit{Initial\_point} together with a probabilistic decision tree. Figure \eqref{fig:qCBR} shows the changes that will be introduced to obtain the qCBR, and figure \eqref{fig:Funct_qCBR} shows the detail of the functional blocks implemented with the specific problem of social workers. The two VQE blocks and the Variational Quantum Classifier are presented before detailing them.

\subsubsection{Variational Quantum Eigensolver}
\begin{figure}[!ht]
\centering
\includegraphics[width=0.4\textwidth]{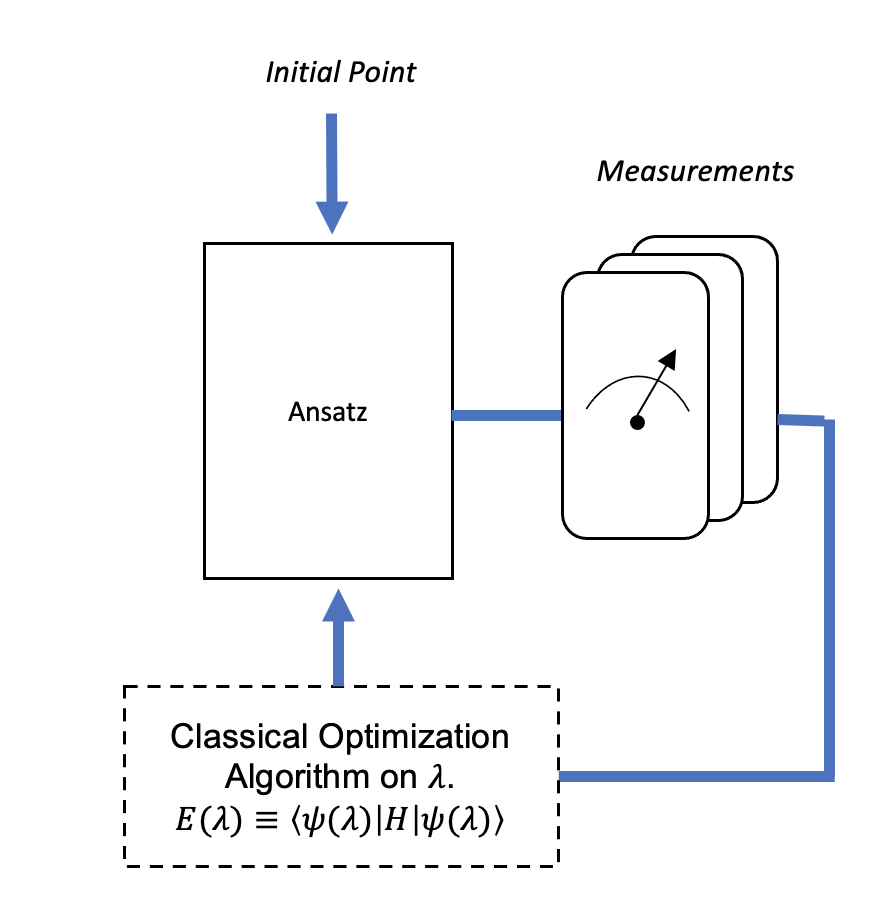}
\caption{We start from an initial position of the initial values of the Ansatz parameters. Normally we start from initial equiprobable values between all qubits. Then, depending on the measured value, a classical and external optimizer is used to feedback the new values of the ansatz parameters. So on until reaching the value of the minimum energy equivalent to the ground state of our Hamiltonian as defined by the variational principle.
If we want to start the optimization from a point other than the initial state, we can load some positions (Initial point) equivalent to some energy values on the ansatz. In the case of qCBR, the key is that having saved the initial point values of each schedule already calculated will save optimization time since it will be possible to start from a fairly optimal point. }
\label{fig:VQE}
\end{figure}

VQE (figure \eqref{fig:VQE}) is a classical hybrid quantum algorithm that combines aspects of quantum mechanics with the classical algorithm, and its objective is to find approximate solutions to combinatorial problems. One of the fundamental approaches is to map combinatorial problems into a physics problem. That is, about a problem that can be formulated in terms of a Hamiltonian Ising model. Therefore, the identification of the solution to the combinatorial problem is linked to finding the ground state of this physics problem. As a result, the goal is to find the ground state of this Hamiltonian.
The unknown eigenvectors are prepared by varying the experimental parameters and calculating the Rayleigh-Ritz ratio \cite{Wu2005} in a classical minimization, figure \eqref{fig:VQE}. At the end of the algorithm, the reconstruction of the eigenvector that is stored in the final set of experimental parameters that define the state would be performed.

From the variational principle, the following equation $\langle H \rangle _{ \psi   \left( \overrightarrow{ \theta } \right) } \geq  \lambda _{i}$ can be reached, with $\lambda _{i}$  as eigenvector and  $\langle H \rangle _{ \psi \left( \overrightarrow{ \theta } \right)}$  as the expected value. This way, the VQE finds \eqref{expectative_value} as an optimal choice of parameters  $\overrightarrow{\theta }$, that the expected value is minimized and that a lower eigenvalue is located.
\begin{equation}
\label{expectative_value}
 \langle H \rangle =\langle\psi\left(\theta  \right)\vert H \vert\psi\left(\theta  \right)\rangle 
\end{equation}
The VQE is used here with and without \textit{Initial\_point}\cite{Qis21} for the synthesis task. In the detailed explanation of the work, more detail will be given. The \textit{Initial\_point} is the starting point (initial parameter values) for the optimizer. Without this starting point, the VQE will search the ansatz for a preferred point, and if not, it will just calculate a random one. This possibility is essentially useful, such as when there are reasons to believe that the outcome position is close to a particular spot. Furthermore, in the qCBR, this \textit{Initial\_point} will be tremendously useful for reshaping the retrieved solution if the latter is not the most optimal.

\subsubsection{Quantum classifiers}
The variational quantum classifier belongs to the variational algorithms like VQE, where classically tunable parameters of a unit circuit are used to minimize the expected value of an observable. The great novelty resides in loading the data in the variational system.

\begin{figure}[!ht]
\centering
\includegraphics[width=0.5\textwidth]{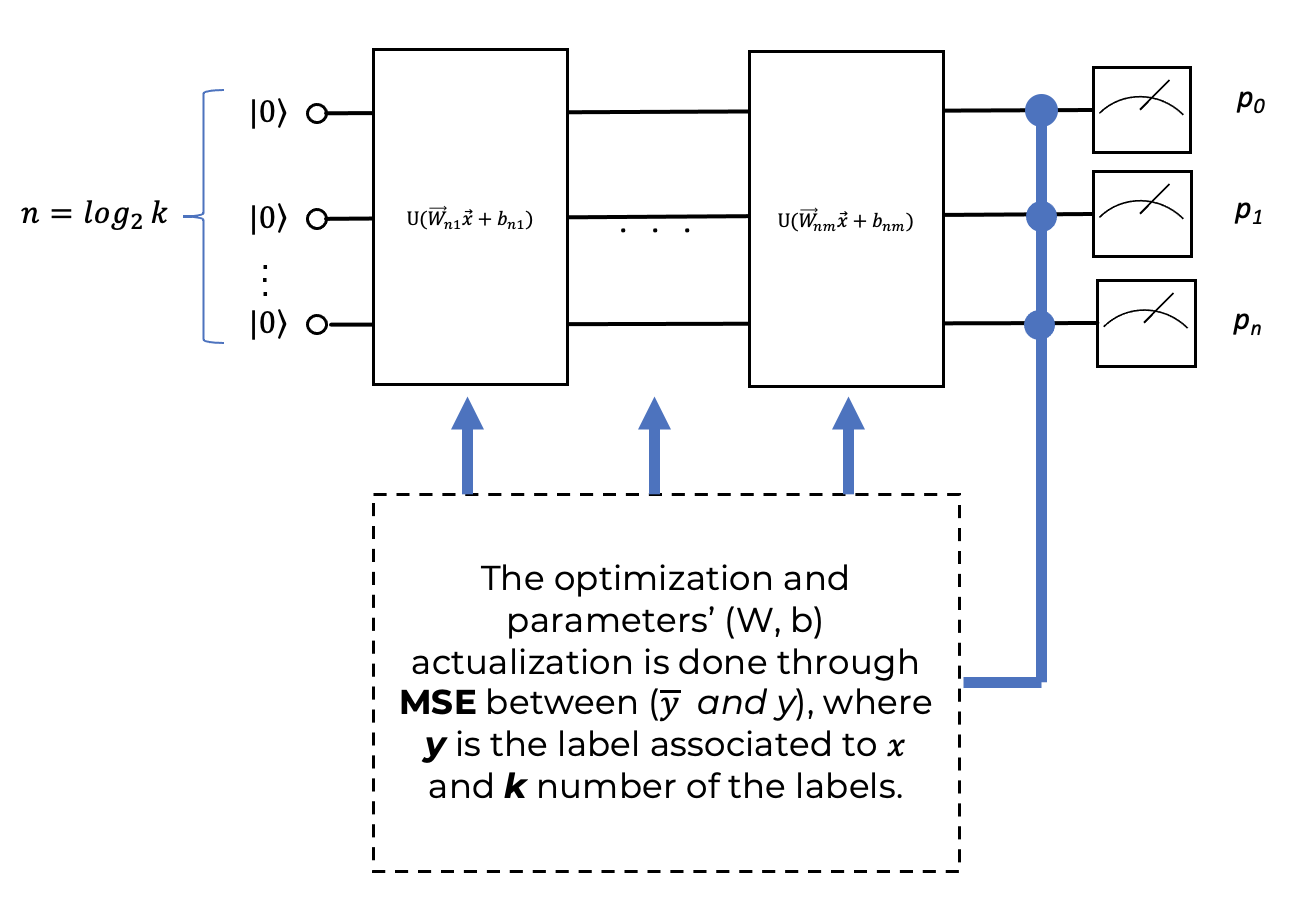}
\caption{This is the variational classifier's diagram block used in the qCBR,  we use the data re-uploading technique to create an n-dimensional classifier as if it were a neural network where the non-linearity of the quantum gates will act as an activation function, and we will use the model $y = Wx + b$.}
\label{fig:class_Gen_}
\end{figure}

We have designed a classifier that emulates neural networks solving the function $Wx + b$, with $W$ and $b$ the parameters and $x$, the sample data to be classified. The non-linearity of the quantum gates is used to implement the activation function $f (Wx + b)$ given $Wx + b$. The figure \eqref{fig:class_Gen_} provides us with the block diagram of the classifier. The optimization and parameters' $(W, b)$ actualization are done in the first step, MSE between ($\bar y$ and $y$), where $y$ is the label associated with $x$ and $k$ of the labels.

\begin{figure*}[t!]
\centering
\includegraphics[width=1\textwidth]{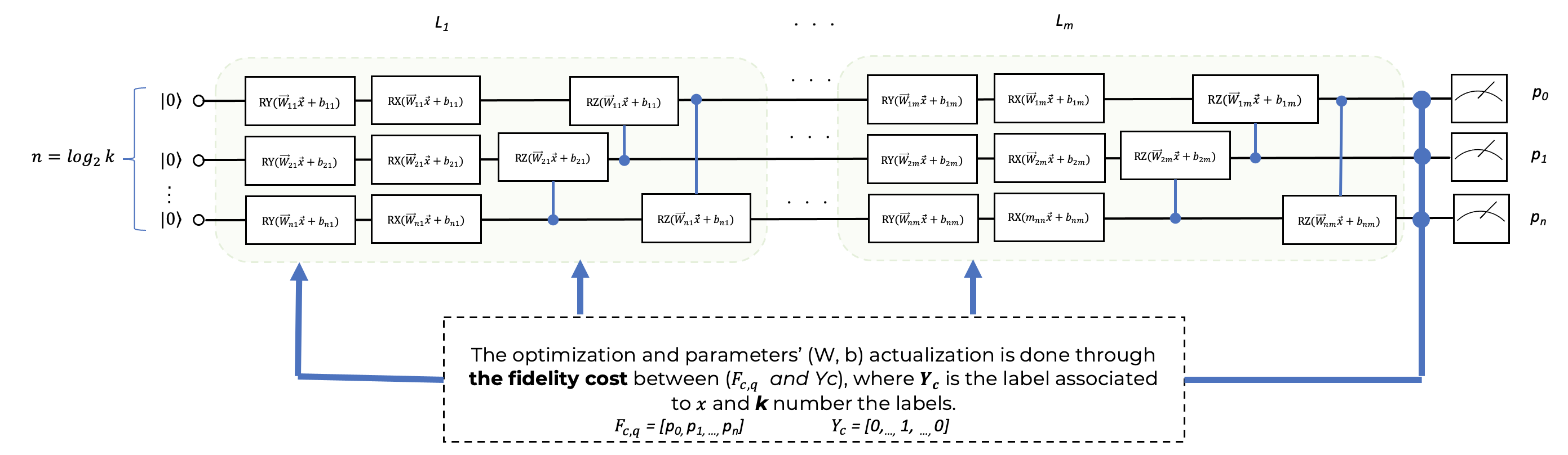}
\caption{This is the design of the classifier implemented in qCBR. Considering that $x$ is the input data of dimension $l$, $Y_{c}$ is the label class of $x$, $k$ is the number of labels, and we use the fidelity cost \eqref{fidelity_cost_function}. It is worth mentioning that the class of the tags $Y_{c}$ and, in this case, coincides with the computational base; thus, we save the target class. In this figure, we got $m$ layers.}
\label{fig:Classi_details}
\end{figure*}

The detailed operations of the classifier are given by the figure \eqref{fig:Classi_details} where the quantum gates, $R_{y}$, $R_{x}$ and $C_{RZ}$ are used to define the block. In our, the optimization and parameters' ($W$, $b$) actualization are done through the fidelity cost between $\alpha _{c,q}F_{c,q} \left( \overrightarrow{ \theta },\overrightarrow{ \omega ,}\overrightarrow{x}_{ \mu } \right)$ and $Y_{c} \left( \overrightarrow{x}_{ \mu } \right)$, where $\overrightarrow{x}_{ \mu }$ are the training points and  $\overrightarrow{ \alpha }= \left(  \alpha _{1}, \ldots , \alpha _{C} \right)$  are introduced as class weights to be optimized together with  $\overrightarrow{ \theta } $,  $ \overrightarrow{ \omega ,}$  are the parameters and  $Q$  the numbers of the qubits. Counting on  $Y_{c} \left( \overrightarrow{x}_{ \mu } \right)$  as the fidelity vector for a perfect classification and $F_{c,q} \left( \overrightarrow{ \theta },\overrightarrow{ \omega ,}\overrightarrow{x}_{ \mu } \right) =\langle \psi _{c} \vert   \rho _{q} \left( \overrightarrow{ \theta },\overrightarrow{ \omega ,}\overrightarrow{x} \right)  \vert   \psi _{c} \rangle$.

Since a universal quantum classifier of $n$ qubits is needed for the purpose of this paper, see figure \eqref{fig:class_Gen_}, a sub-base in the Hilbert vector space of equitably dividing the hyperplane $Z$ is described as follows.
Let  $ B= \left(\{ i,j,k,l,m,n,o,p  \}\right)$ be a sub-base within the Hilbert vector space, for the space of the classes  $C^{2^{q}}$, the coordinates of the target classes are defined by expression \eqref{measurement_basis} with $q$ as the number of the qubits.
\begin{equation}
\label{measurement_basis}
\begin{aligned}
&\{i \left( \text{1,0,0,0,0,0,0,0} \right)  ;j \left( \text{0,1,0,0,0,0,0,0} \right) ;k \left( \text{0,0,1,0,0,0,0,0} \right) ; \\
&l \left( \text{0,0,0,1,0,0,0,0} \right) ; m \left( \text{0,0,0,0,1,0,0,0} \right) ;n \left( \text{0,0,0,0,0,1,0,0} \right) ; \\
&o \left( \text{0,0,0,0,0,0,1,0} \right) ;p \left( \text{0,0,0,0,0,0,0,1} \right) \}
\end{aligned}
\end{equation}

The Ansatz design and data loading (variables  ${x_i}$ similar to neural networks)\cite{UAT2021} are given by equation \eqref{Ansatz}, and its analysis is detailed in \ref{sec:App-Ansatz}.
\begin{equation}
\label{Ansatz}
U=\left(\theta ,{x} \right)= R_{x}\left( { \theta _{1}}{x}+{ \theta _{2}} \right) R_{z} \left( { \theta _{3}}\right) 
\end{equation}

\subsubsection{Memory Structure}
\begin{figure}[!ht]
\centering
\includegraphics[width=0.45\textwidth]{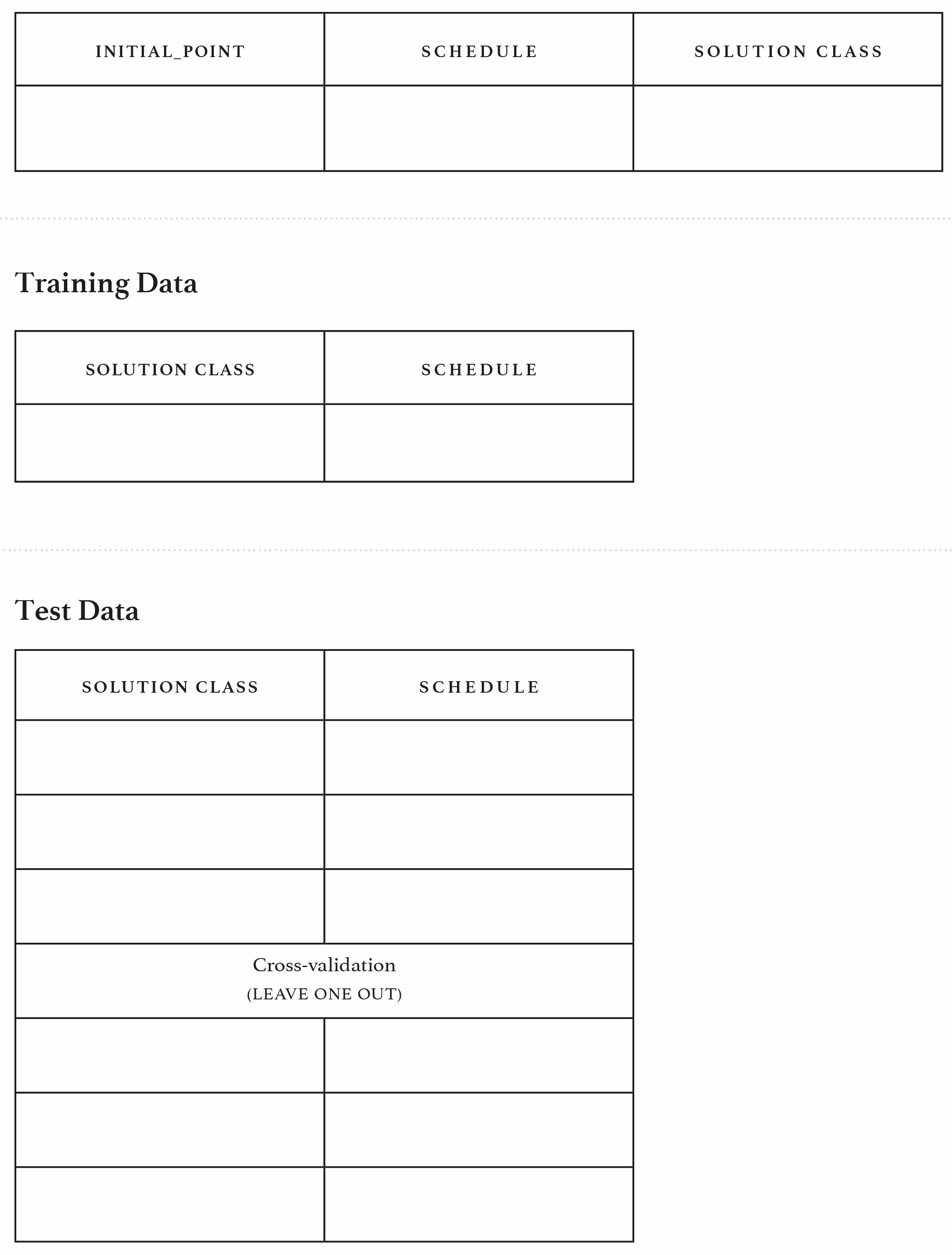}
\caption{qCBR's Memory structure, based on classical RAM. The use the cross-validation technique helps to improve the quality of the classifier training \cite{Mic}.}
\label{fig:qMem}
\end{figure}

\begin{figure}[!ht]
\centering
\includegraphics[width=0.45\textwidth]{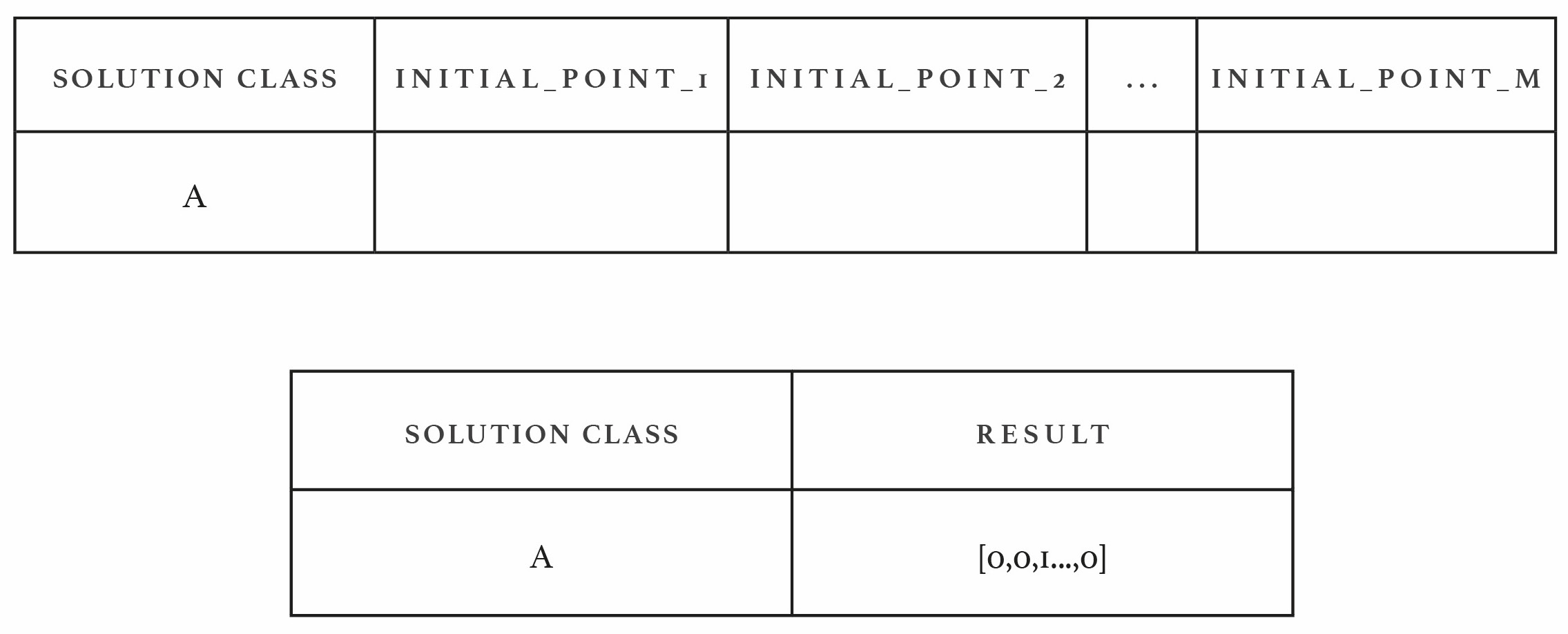}
\caption{Memory structure used for the Training System, based on classical RAM. A label identifies each class, and in this case, $A$ is one of the labels. For more detail, refer to the code \cite{Ade21}.}
\label{fig:mem1}
\end{figure}

Next, some test benches based on the memory structure described in figures \eqref{fig:qMem} and \eqref{fig:mem1} are defined to train the parameterized quantum circuit, and its performance is analyzed in terms of the circuit architecture. The results, discussions and annexe sessions will emphasize the classifier with or without entanglement and a comparative study with different ansatzes.
\newline

The memory structure of the qCBR' retention system is given in figure \eqref{fig:qMem}. The solution class (target) corresponds to the paths each Social Worker will take between the different patients for a specific schedule, representing these paths as an adjacency matrix, such as:

\begingroup
\renewcommand*{\arraystretch}{0.7}
\begin{equation}
\label{eq:SWAdjacencyMatrixExample}
 SOL_{SWP} = \begin{pmatrix}
    0 & x_{0,1} & x_{0,2}\\
    x_{1,0} & 0 & x_{1,2}\\
    x_{2,0} & x_{2,1} & 0\\
 \end{pmatrix}
\end{equation}
\endgroup

where $x_{i,j}$ is a binary variable, the rows of the matrix represent the origin node and the column the destination node of the path.\\

Each solution class is represented as a label (e.g., 'A') and is related to the different initial points associated with each of the samples that make up the training dataset can be seen. This solution class is also associated with the result of the VQE. 

\subsection{The qCBR solving the Social Workers' Problem}
A real problem already developed in these references \cite{Atc20}\cite{Atc202} is used to correctly test the proposed and implemented qCBR.
The social workers' schedule problem (SWP) is defined by generating an optimal visiting schedule for the social workers, who visit their patients at home, to provide them with personalized attention and assistance depending on the patient's pathology. More details about the SWP can be found at \cite{Atc20}.
However, and for the better understanding of this article, let us recall the simplified objective function subjected to the restrictions in Hamiltonian form for the SWP as follows:

\begin{equation}
\label{SWP_Formulation}
\begin{aligned}
& H= \sum _{ij \in E}^{} \left( d_{ij}+ \varepsilon \frac{ \left(  \tau_{i-} \tau_{j} \right) ^{2}}{d_{\max }-d_{\min }} \right) x_{i,j} \\
& + A \sum _{i=1}^{n} \left( 1- \sum _{j \in  \delta  \left( i \right) ^{+}}^{N}x_{i,j} \right) ^{2} +A \sum _{i=1}^{n} \left( 1- \sum _{j \in  \delta  \left( i \right) ^{-}}^{N}x_{ji} \right) ^{2} \\
& +A \left( k- \sum _{i \in  \delta  \left( 0 \right) ^{+}}^{N}x_{0,i} \right) ^{2} +A \left( k- \sum _{j \in  \delta  \left( 0 \right) ^{+}}^{N}x_{j,0} \right) ^{2}
\end{aligned}
\end{equation}

Where  $A$ is the Lagrange multiplier which is a free parameter such that $A > max\left( d_{ij}+ \varepsilon \frac{ \left(  \tau_{i-} \tau_{j} \right) ^{2}}{d_{\max }-d_{\min }} \right)$, where  $x_{ij}$  are the decision and binary variables of the paths between two patients,\   $d_{ij}$  is the distance between the patient $i$ and the next $ ~j$ \ and   $g_{ij}= \left( d_{ij}+ \varepsilon \frac{ \left(  \tau_{i-} \tau_{j} \right) ^{2}}{d_{\max }-d_{\min }} \right)$  is the non-negative time window's function and it is mapped on a quadratic function to weigh extremal distances (shortest concerning the greatest ones). Let us consider that the initial weight function  $w_{ij}=d_{ij}$ is a distance function because one wants to make $ g_{ij}$  behave like  $d_{ij}$, and thus be able to take full advantage of the initial objective function's behaviour.

Let  $\varepsilon$ be positive and represent a weighted degree parameter of the time window function;  $\tau_{i}$ is the starting worker time of a slot of time for patient $~i~$ and  $\tau_{j}$ for the patient $~j$. With  $d_{\max }$  as the maximum distance between all patients and  $d_{\min }$ the minimum one. Hence, let us define the non-negative time windows $T_{ij}= \left(  \tau_{i-} \tau_{j} \right) >0 $.

\begin{figure}[!ht]
\centering
\includegraphics[width=0.45\textwidth]{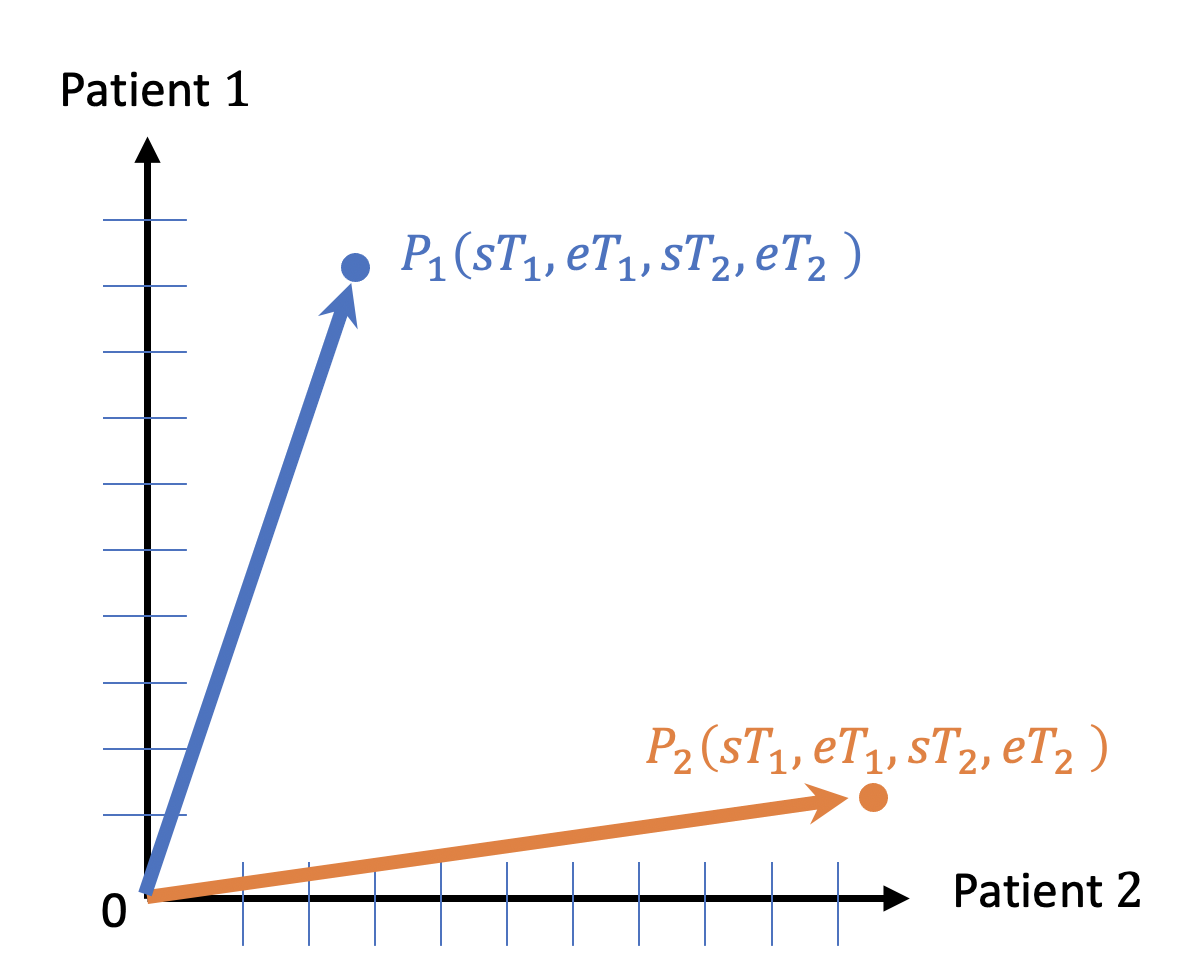}
\caption{The SWP represented in vector form to take advantage of the Hilbert vector space's characteristics within quantum computing. It is seen that each patient represents a dimension and the points that make up the dataset a dimension of $2n$ coordinates, $n$ being the total number of patients. In this figure, to simplify the understanding, we use two patients, therefore, two dimensions.}
\label{fig:SWP}
\end{figure}

To fill out the data structure created in the classifier to train and test its predictions, the parameters of the \textit{Initial\_point} obtained by VQE are abstracted from the result. Furthermore, it is composed of each class's coordinates with the following parameters: start time $sT$ and end time  $eT$  of patient  $1$ to  $n$, where  $n$  is the maximum number of patients in the app. Finally, figure \eqref{fig:SWP} summarizes the data's representation and description that make up the training dataset.

Each coordinate's class corresponds to the VQE solution following the memory structure in figure \eqref{fig:qMem}. Where the class number of the classifier is given by equation \eqref{Num_Sol_SWP} taking into account the conditions that every worker has a patient and that the workers are indistinguishable (that is, it doesn't matter whether the social worker $m_{1}$ takes care of the patient  $n_{1}$  and  $m_{2}$  takes care of $n_{2}$  or vice versa).
\begin{equation}
\label{Num_Sol_SWP}
    N_{SOL_{SWP}}=\frac{1}{m!} \sum_{k=0}^{m-1}(-1)^{k}{m \choose m-k}(m-k)^{n}
\end{equation}

With $ n $ the number of patients, $ m $ the number of social workers and, knowing that the appearance of patients is ordered in the schedule (from oldest to most recent), $ n_{1} $ will be the patient with the first schedule and $ n_{k} $ the patient with the last one.

In this article all the tests done are for  $n = 4$  with the data structure equal to  $ ( sT_{1},eT_{1},sT_{2},eT_{2},sT_{3},eT_{3},sT_{4},eT_{4})$; an 8-dimensional vector for each social worker visit the patient. In this case, the number of qubits will be defined by  $q=\log_{2}(N_{SOL_{SWP}})=3$. These qubits are used to instance the quantum classifier, and it is worth to mention that the classifier must have  $N_{SOL_{SWP}}$ classes.

The detail of the qCBR’s implementation and analysis is in the Appendix \ref{sec:App-CBR}.

\begin{figure*}[t!]
\centering
\includegraphics[width=1\textwidth]{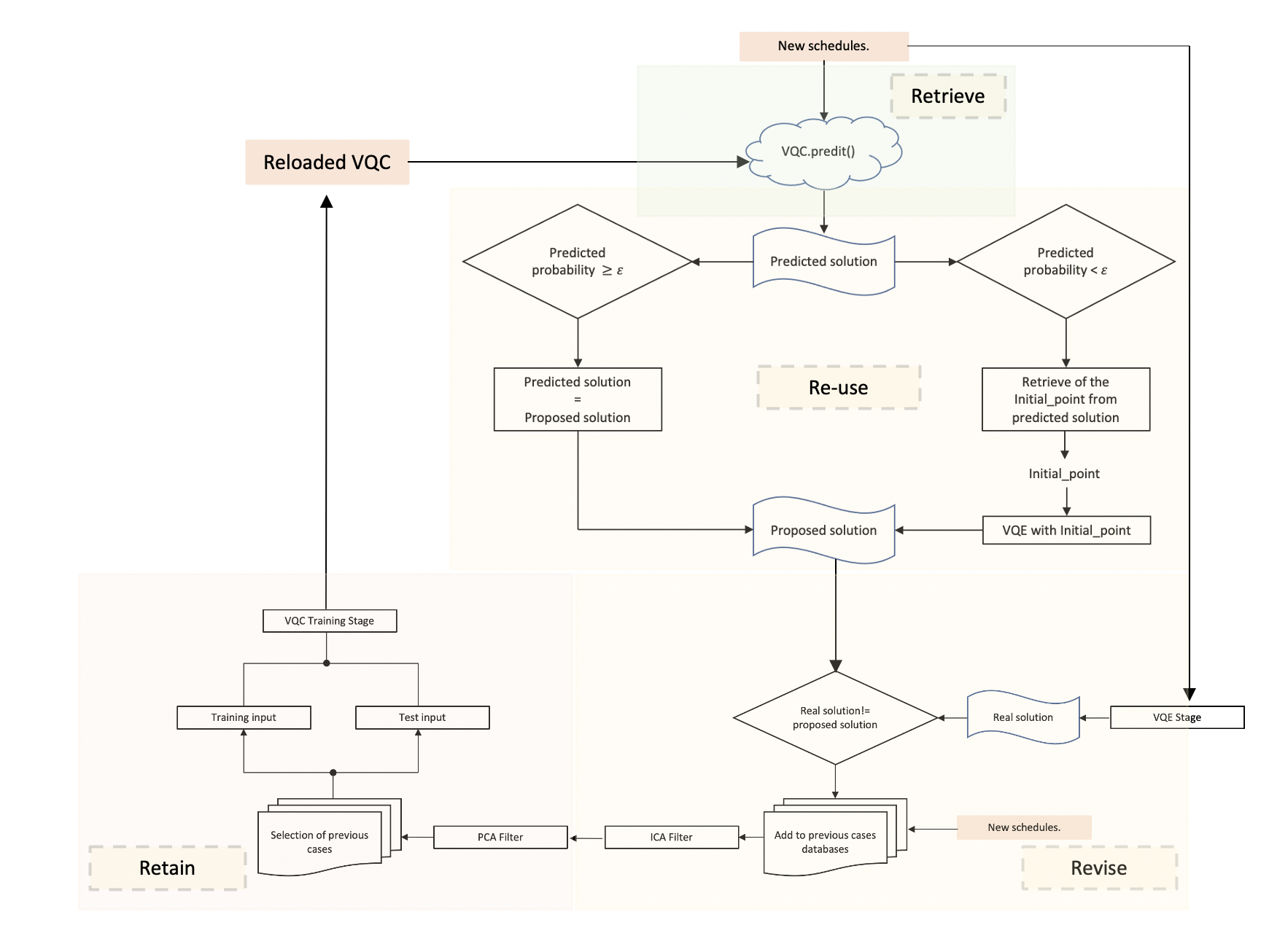}
\caption{Block diagram proposed for the resolution of qCBR, considering a real dataset with an overlap problem between the data components. This block diagram contemplates the treatment of the input data and the use of ICA\cite{Hyv01_interscience} and PCA\cite{Jon14} before training and classifying the data. In this version, a classifier based on the re-uploader has been designed to be in charge of the classification tasks. And for the synthesis tasks, a decision tree passed in the classifier predictions have been used together with the VQE plus the \textit{Initial\_point}.}
\label{fig:Funct_qCBR}
\end{figure*}

\section{Results}\label{sec:result}
When testing the classifier, a section of the sample database, schedules previously solved by the VQE algorithm to obtain its corresponding true solution (\textit{ground truth}), was used as test samples (applying “Leave-one-out” cross-validation). Then, the total accuracy of the classifier predictions was obtained based on the ratio between the number of labels predicted correctly and the total number of labels.

\begin{figure*}[t!]
\centering
\includegraphics[height=6cm]{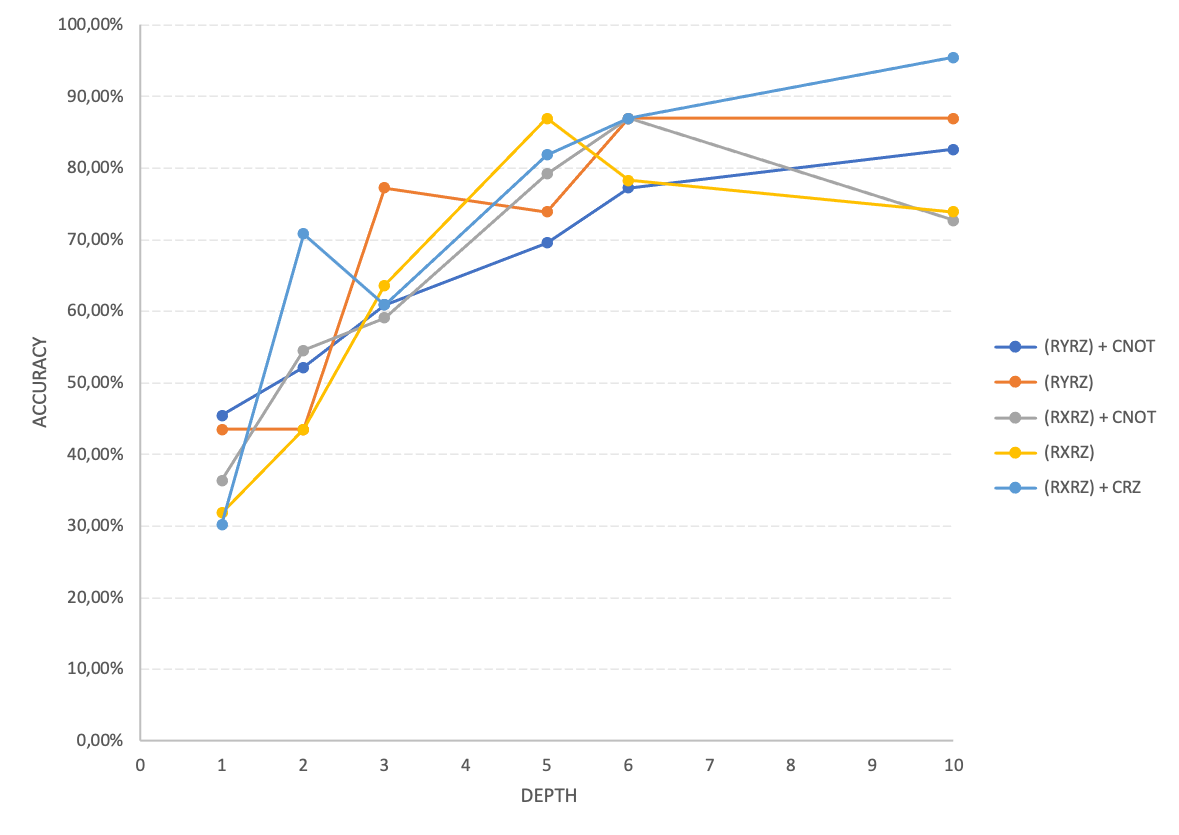}
\includegraphics[height=6cm]{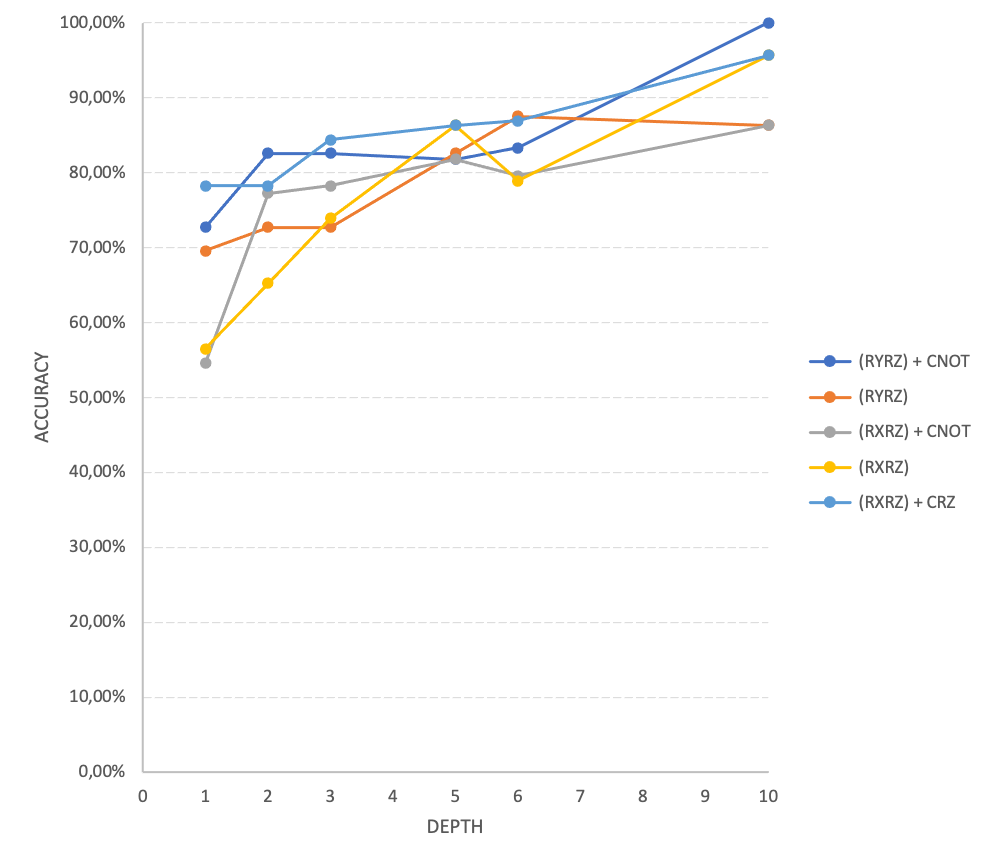}
\caption{Comparative graphs between different ansatzes, taking into account the classifier's accuracy as a function of depth. Represents the evolution of the ansatzes of dimension two and eight.}
\label{fig:Bench_C}
\end{figure*}

\begin{figure}[t!]
\centering
\includegraphics[width=.49\textwidth]{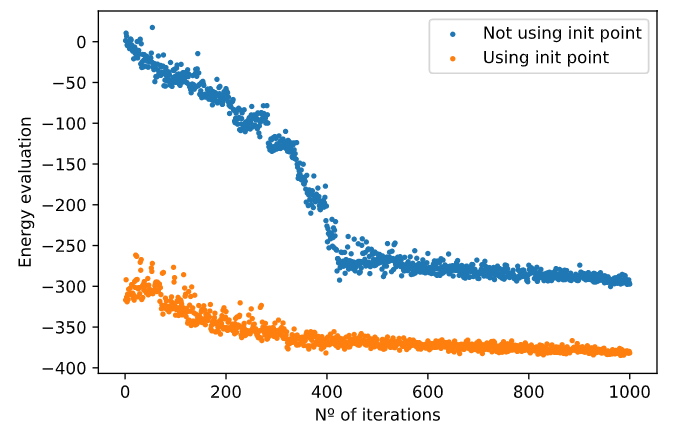}
\caption{Energy comparison between VQE algorithm without using or using the initial point. It is noticeable how the first one tends to stabilize after multiple iterations (approximately 400), starting the search for a minimum from a random starting point (depending on the seed provided). Meanwhile, the second one is capable of stabilizing and reach a solution close to the absolute minimum with much less iterations, starting from energy point evaluation close to the real energy solution}
\label{fig:Bench_VQE_Init}
\end{figure}

Figures \eqref{fig:Bench_C} to \eqref{fig:Bench_VQE_Init} show the implementation outcomes performed in \textit{qibo}\cite{qibo} and \textit{qiskit}\cite{mckay2018qiskit,Qis21} to identify the best model architecture and represent functions similar to qCBR.

Tables \eqref{tab:results_qCBR_SW_Full} to \eqref{tab:results_qCBR_SW_4x3SW} show the global results of qCBR solving the SWP. In table \eqref{tab:results_qCBR_SW_Full}, the outcome of the different tested scenarios can be observed. Varying the number of patients, social workers, and the quantum circuit's depth to see the global hit number of the qCBR. In table \eqref{tab:results_qCBR_5x4SW}, we can observe the resolution of the SWP, considering five patients, four social workers and setting the depth of the quantum circuit to eight. Through this scenario, the behaviour of the qCBR can be observed considering the number of cases carried out. It can be seen how the system begins to give more than satisfactory results after exceeding the threshold of the 240 results stored in the case memory.
Table \eqref{tab:results_qCBR_SW_4x3SW} repeats the steps of table \eqref{tab:results_qCBR_5x4SW} with the only change of the input data; the number of patients and social workers.
Tables \eqref{tab:results_CBR_SW_5x4SW} to \eqref{tab:results_CBR_KNN_SW_Full} show the result of the implementation of the classical CBR leveraged on ANN and KNN to solve the SWP.

Tables \eqref{tab:results_qCBR_5x4SW} and \eqref{tab:results_qCBR_SW_4x3SW} represent the outcomes of the qCBR and show better results than the ones obtained with the classical CBR (tables \eqref{tab:results_CBR_SW_5x4SW} and \eqref{tab:results_CBR_SW_4x3SW}). Tables \eqref{tab:results_CBR_KNN_SW_Full}, \eqref{tab:results_CBR_NN_SW_Full} and \eqref{tab:results_qCBR_SW_Full} show the degree of scalability of the qCBR as a function of the variation in the number of patients and social workers. It has also been seen that qCBR is much better shared with overlapping as we wanted to demonstrate.

\begin{table}[t!]
\centering
\begin{tabular}{ |c|c|c|c|c|c|  }
 \hline
 \multicolumn{6}{|c|}{ qCBR solving the Social Workers' Problem } \\
 \hline
 \#Patients & \#SW & \#Qubits &\#Layers  & \#Cases & Accuracy\\
 \hline
 3   & 2 & 6  &  2  & 580 & 82.5\\
 4   & 3 & 12 &  3  & 580 & 82\\
 5   & 2 & 20 &  4  & 580 & 82.5 \\
 5   & 3 & 20 &  5  & 580 & 87\\
 5   & 4 & 20 &  8  & 580 & 92.8\\
 5   & 4 & 20 &  10 & 580 & 100\\
 \hline
\end{tabular}
\caption{The result of the qCBR with a variational classifier and using the VQE and the Initial\_point with some decision trees as a synthesizer \eqref{fig:Funct_qCBR}. This table shows the different studies made as a function of the quantum circuit's depth (layer). number of the patients and the social workers. The accuracy of the classifier is the maximum with the number of layers equal to 10. SW denotes Social Workers.}
\label{tab:results_qCBR_SW_Full}
\end{table}

\begin{table}[t!]
\centering
\begin{tabular}{|c|c|c|  }
 \hline
 \multicolumn{3}{|c|}{ qCBR solving the Social Workers' Problem } \\
 \multicolumn{3}{|c|}{ For 5 patients and 4 socials workers} \\
 \hline
 Layers &\#Cases & Accuracy\\
 \hline

& 20  & -  \\ 
&50  & 12.5\\
&100 & 72.5\\
8 &240 & 92.1\\ 
&340 & 95.5\\
&480 & 97.2\\
&500 & 98.7 \\
&580 & 99.1\\
 \hline
\end{tabular}
\caption{The result of the qCBR for a number of patients and social workers fixed at 5 and 4, respectively. The better behaviour of qCBR can be observed for some cases greater than 240. To have a good functioning of the qCBR, it must be iterated with the social workers' dataset one 239 times. And at the case number of 240, we will have an accuracy of 92\%. The "accuracy" value is the percentage of the number of correct solutions found by the qCBR.
A hyphen (-) denotes that no solution was found within the 20 cases. All these tests were done for the quantum circuit depth (layers) equal to 8.
}
\label{tab:results_qCBR_5x4SW}
\end{table}

\begin{table}[t!]
\centering
\begin{tabular}{|c|c|c|}
 \hline
 \multicolumn{3}{|c|}{ qCBR solving the Social Workers' Problem } \\
 \multicolumn{3}{|c|}{ For 4 patients and 3 socials workers} \\
 \hline
 Layers &\#Cases & Accuracy\\
 \hline

& 20  & -  \\ 
&50  & 11.5\\
&100 & 73.1\\
8 &240 & 91.1\\ 
&340 & 91.9\\
&480 & 96.6\\
&500 & 98.1 \\
&580 & 99.0\\
 \hline
\end{tabular}
\caption{The result of the qCBR for a number of patients and social workers fixed at 4 and 3, respectively. The better behaviour of qCBR can be observed for some cases greater than 240. To have a good functioning of the qCBR, it must be iterated with the social workers' dataset one 239 times. And at the case number of 240, we will have an accuracy of 91\%. The "accuracy" value is the percentage of the number of correct solutions found by the qCBR.
A hyphen (-) denotes that no solution was found within the 20 cases. All these tests were done for the quantum circuit depth (layers) equal to 8.
}
\label{tab:results_qCBR_SW_4x3SW}
\end{table}

\begin{table}[t!]
\centering
\begin{tabular}{|c|c|c|}
 \hline
 \multicolumn{3}{|c|}{ CBR with KNN solving the Social Workers' Problem } \\
 \multicolumn{3}{|c|}{ For 5 patients and 4 socials workers} \\
 \hline
 Layers &\#Cases & Accuracy\\
 \hline
& 20  & -  \\ 
&50  & 42.9\\
&100 & 46.5\\
1 &240 & 52.6\\ 
&340 & 55.3\\
&480 & 56.8\\
&500 & 60.7 \\
&580 & 63.1\\

 \hline
\end{tabular}
\caption{The classical CBR result on KNN classifier for a number of patients and social workers fixed at 5 and 4, respectively. The better behaviour of this CBR can be observed for some cases greater than 240. To have a good functioning of the CBR, it must be iterated with the social workers' dataset one 239 times. And at the case number of 240, we will have an accuracy of 52.6\%. The "accuracy" value is the percentage of the number of correct solutions found by the CBR leveraged on KNN, applying a 10-KFold cross-validation process.
A hyphen (-) denotes that no solution was found within the 20 cases. All these tests were done for the layer equal to 1.}
\label{tab:results_CBR_SW_5x4SW}
\end{table}

\begin{table}[t!]
\centering
\begin{tabular}{|c|c|c|}
 \hline
 \multicolumn{3}{|c|}{ CBR with KNN solving the Social Workers'Problem } \\
 \multicolumn{3}{|c|}{ For 4 patients and 3 socials workers} \\
 \hline
 Layers &\#Cases & Accuracy\\
 \hline
& 20  & -  \\ 
&50  & 55.1\\
&100 & 58.5\\
1 &240 & 70.3\\ 
&340 & 71.1\\
&480 & 73.6\\
&500 & 74.8 \\
&580 & 76.8\\

 \hline
\end{tabular}
\caption{The classical CBR result on KNN classifier for a number of patients and social workers fixed at 4 and 3, respectively. The better behaviour of this CBR can be observed for some cases greater than 100. To have a good functioning of the CBR, it must be iterated with the social workers' dataset one 239 times. And at the case number of 240, we will have an accuracy of 70.3\%. The "accuracy" value is the percentage of the number of correct solutions found by the CBR leveraged on KNN, applying a 10-KFold cross-validation process.
A hyphen (-) denotes that no solution was found within the 20 cases. All these tests were done for the layer equal to 1.}
\label{tab:results_CBR_SW_4x3SW}
\end{table}

\begin{table}[t!]
\centering
\begin{tabular}{ |c|c|c|c|c|  }
 \hline
 \multicolumn{5}{|c|}{  CBR leveraged by CNN solving the Social Workers' Problem } \\
 \hline
 \#Patients & \#SW  &\#Layers  & \#Cases & Accuracy\\
 \hline
 3   & 2 &   2  & 580 & 65.4\\
 4   & 3 &   2  & 580 & 43.3\\
 5   & 2 &  2  & 580 & 37.3 \\
 5   & 3 &  2  & 580 & 26.3\\
 5   & 4 &  2 & 580 & 45.2\\

 \hline
\end{tabular}
\caption{CBR with a neural network classifier and a backtracking algorithm as a synthesizer. SW denotes Social Workers.}
\label{tab:results_CBR_NN_SW_Full}
\end{table}

\begin{table}[t!]
\centering
\begin{tabular}{ |c|c|c|c|c|  }
 \hline
 \multicolumn{5}{|c|}{CBR with KNN solving the Social Workers' Problem } \\
 \hline
 \#Patients & \#SW  &\#Layers  & \#Cases & Accuracy\\
 \hline
 3   & 2 &  1  & 580 & 95.6\\
 4   & 3 &  1  & 580 & 77.8\\
 5   & 2 &  1  & 580 & 47.8 \\
 5   & 3 &  1  & 580 & 44.7\\
 5   & 4 &  1 & 580 & 63.1\\
 \hline
\end{tabular}
\caption{CBR with a KNN classifier and a backtracking algorithm as a synthesizer. SW denotes Social Workers.}
\label{tab:results_CBR_KNN_SW_Full}
\end{table}

Also, we experimented by skipping the Principal Component Analysis (PCA) module \cite{Jon14}\cite{Ewi19}, Independent Component Analysis (ICA) \cite{Hyv01_interscience} and creating a classifier of the same dimension as the data (8 dimensions). The results obtained have been very satisfactory at the Ansatz's accuracy and depth level. Still, the need to change the \textit{BFGS} \cite{BFGS_Limted} optimizer to the SPSA \cite{Jam01} has become visible due to its slow convergence for the number of data and high parameters. Figure \eqref{fig:Bench_C} describes the behaviour and compare the two scenarios.

Later the Re-use module was analyzed using VQE with \textit{Initial\_point} to synthesize the predicted results. In the graph shown in figure \eqref{fig:Bench_VQE_Init}, it is observed how the algorithm, without initial parameters, tends to use a high energy constant of variation to quickly reach an approximation of the fundamental state. Which makes it have to progressively, after several iterations,  $n$ , reduce said constant to find the local minimum. On the other hand, when using an \textit{Initial\_point}, the algorithm does not need to start with a high variation to reach energy bands close to the ground state since it is much closer to said energy, reducing the number of iterations necessary reach to the local minimum. We can then see how qCBR can afford to run VQE with \textit{Initial\_point} to refine the accuracy of its results since it requires fewer iterations to find the solution closest to the minimum, not assuming such a high computational cost as it would be running VQE without initial parameters.

\begin{table}[t!]
    \centering
    \begin{tabular}{c|c}
         \textbf{Methods}   & \textbf{Complexity} \\
         \hline
         Retrieve & $O(log NM)$  \\
         Re-use &  $ O(Klog(N)+log NM)$ \\
         Revise &  $O(log(N))+ICA)$ \\
         Retain &  $O(log NM)+PCA$ \\
    \end{tabular}
    \caption{Table of the complexity of qCBR counting the PCA cases and the ICA complexity. In this case, K, the number of shots of the VQE, is fixed to 50.}
    \label{tab:qCBR_complexity}
\end{table}

The qCBR complexity (Table \eqref{tab:qCBR_complexity}) is provided below where it can be seen that the \textit{Retain} is the highest cost operation and has an exponential improvement compared to a \textit{Retain} of a classic CBR that is usually of the order of $O( M^{2}(M+N))$ \cite{Pat14}

\section{Discussions}\label{sec:Discussions}
Firstly, the proposed qCBR works very well and meets the objectives set using quantum computing to create efficient quantum Case-Based Reasoning. One of the issues to comment on is the improvement observed in figure \eqref{fig:Bench_C}  with respect to the 2 and 8-dimensional classifiers. Due to the small number of depths, but with many more parameters, the 8-dimensional classifiers have an average of about 25$\%$  of improvements over the 2-dimensional ones. With this result, in the case of not wanting an accuracy of around 95$\%$, shallow depth could be used, and computation time saved, depending on the problems.
Despite all these improvements, it is essential to highlight some aspects to refine. In the intelligent system that allows deciding the proposed solution, now, the average of the \textit{Initial\_point} of each solution class samples' \textit{Initial\_point} is used. It could still be seen based on the predicted solution, which \textit{Initial\_point} is the most suitable for the solution to propose. Thus, the cases to be re-used could be better classified.

Also, one of the improvements is to train the classifier with noisy data further so that the qCBR can adapt to real past situations that adjust to the new situation. Because, in practice, there is usually no past case strictly the same as a new one.

The last improvement is to generalize the qCBR to serve various types of problems (betting problem, financial, software maintenance, human reasoning, etc.). To get it, we must focus on designing the memory of the cases so that different data sizes can be indexed and train the classifier with several other data models. 

Secondly, both QIR \cite{lebedev2020introductory} and qCBR work with a data representation model based on a multidimensional vector in Hilbert space.

This offers the possibility for quantum algorithms to perform a clustering or discrimination of the data within this vector space.

The QIR analyses whether a certain entry is related to other types of documents previously studied and how the classic NLP techniques are performed \cite{chowdhury2003natural, liddy2001natural}. To do this, it projects the input vector introduced concerning the bases of the clusters built corresponding to each class with similar patterns.

At the same time, qCBR follows a similar process for predicting whether an input vector corresponds to a previously analysed class and calculates the probability that each type corresponds to the new vector from the proximity of each vector subspaces generated from each category.

The text representation is transformed to a numeric vector from a process called word2vec \cite{goldberg2014word2vec, rong2014word2vec, church2017word2vec} and doc2vec \cite{lau2016empirical, kim2019multi}, and once the vector is obtained, the process to follow is identical to the one to follow by qCBR. In many cases, seeing references \cite{khrennikov2019quantum, bruza2006quantum, lund1996producing, deerwester1990indexing}, QIR and NLP already predefine the classes to be analysed, either Pop, Rock, etc. By predefining that each axis of the Hilbert space corresponds to a type, this process is similar to the qCBR but without the synthesiser's ability.

The clustering process allows the algorithm to create classes and related documents without specifying the categories; therefore, in the case of QIR, it does not move away from an abstraction of the classical problem of "bag-of-words" parsers of spam.

The creation of the SWP vector subspace over the Hilbert vector space is similar in the references \cite{piwowarski2010exploring, Piwowarski2010} where the authors focus on filters, request and document retrieval.

It is worth noting that the qCBR does not present a barren plateau problem due to the low numbers of qubits, shallow quantum circuit and because we have used local cost functions as advocated by the barren plateau theorem \cite{cerezo2021cost}.

\section{Conclusions and further work}\label{sec:Conclusions}
We observed the outstanding performance of qCBR compared to its classical counterpart on the average accuracy, scalability and tolerance to an overlapping dataset.
Some of the problems of standard and classical CBR have been mitigated in this work. With the design that has been proposed in this work, it has been possible to measure situations of difficult similarity between cases. Despite the non-linear and overlapping attributes, the classifier has been endowed with characteristics that serve to arrive at two similar topics that may seem quite different by having different values in features, but not very important. In the VQE with \textit{Initial\_point}, we can have different \textit{Initial\_point} associated with each training class sample with the same class. With the technique of the average of the "\textit{Initial\_point}", it is possible to solve this problem by providing the qCBR to distinguish the similarity between cases. Another issue that qCBR mostly solves is the time required to classify a new topic.

With the results of the two implementations (classical and quantum CBR), it is observed that the classical CBR designed with the KNN behaves better for some determined cases (table \eqref{tab:results_CBR_KNN_SW_Full}). It is seen that the system has not finished learning thoroughly (table \eqref{tab:results_CBR_SW_5x4SW} and \eqref{tab:results_CBR_SW_4x3SW}) contrary to the qCBR (table\eqref{tab:results_qCBR_5x4SW} and \eqref{tab:results_qCBR_SW_4x3SW}). This is due to its classifier's accuracy, without forgetting the significant contribution of its synthesis system.

Another improvement that qCBR introduces is when retaining cases, implementing a retention system that maintains model cases and that, together, synthesize the real and most important information.
One of the improvements to consider is the implementation of quantum ICA. In this way, the classical ICA analysis's complexity cost will be significantly reduced. Also counting that the PCA is saved since we have an 8-dimensional classifier, the complexity of the qCBR would be that of the classifier plus some setup constants. 

The other exciting line of the future is to design the memory of cases using the quantum technique of random-access memory (qRAM) \cite {qRAM_} to improve the memory of stored cases.

\acknowledgments
The authors greatly thank the IBMQ team, mainly Steve Wood. P.A. thanks Jennifer Ramírez Molino, the Qibo team, Adrian Perez-Salinas and Guillermo Alonso Alonso de Linaje for his support and comments on the manuscript.
\\

\textbf{Compliance with Ethics Guidelines}\\
\\
Conflict of interest: P. Atchade Adelomou, D. Casado Faulí, E. Golobardes Ribé and X. Vilasís-Cardona state that there are no conflicts of interest.

\appendix \label{sec:Appendix}

\section{Variational Quantum Classifier}\label{sec:App-VQC_Reup}
To date, two dominant categories allow to design quantum classifiers. Although almost all are inspired by the classical classifiers (kernel or neural networks) \cite{Abd15}, there is a new category of classifiers that respond to the current era of quantum computing (NISQ); hybrid and variational classifiers.

\subsection{The Ansatz}\label{sec:App-Ansatz}
The Ansatz design inherited from previous works \cite{Atc201,Atc20} \cite{Suk191}. The way to load the data into the Ansatz is inspired by \cite{UAT2021} where the data (variable  $x$) is entered using the weights and biases scheme. In this case, the single-qubit gate that serves as the building block for all Ansatz is given by \eqref{anstaz_1D} similar to neural networks.
\begin{equation}
\label{anstaz_1D}
U= \left(  \theta ,x \right) =R_{x} \left(  \theta _{1}x+ \theta _{2} \right) R_{z} \left(  \theta _{3} \right) 
\end{equation}

\begin{figure}[ht!]	
\includegraphics[width=.3\textwidth]{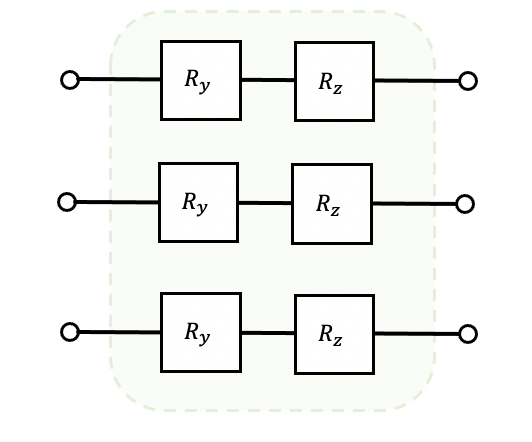}
\caption{$R_y$ and $R_z$ Ansatzes without entanglament used in qCBR experimentation.}
\label{fig:Ansatz_1}
\end{figure}

\begin{figure}[ht!]
\includegraphics[width=.45\textwidth]{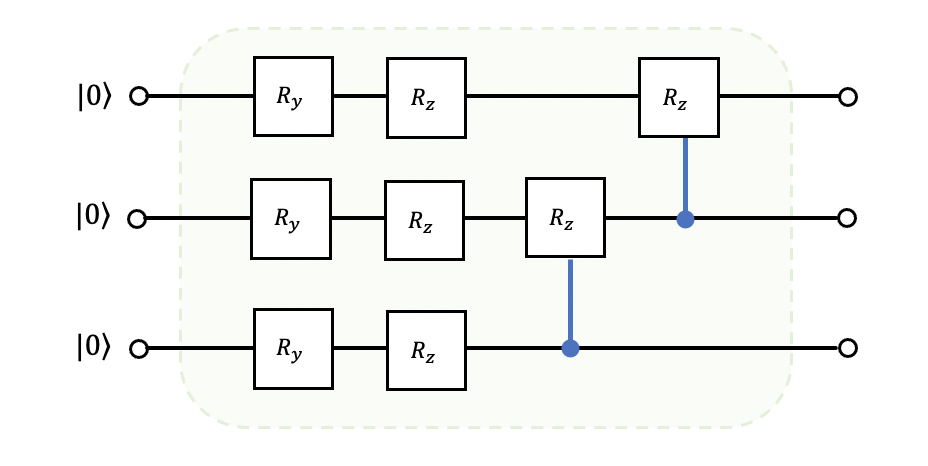}
\caption{$R_y$ and $R_z$ Ansatzes with $CRZ$ entanglement used in qCBR experimentation.}
\label{fig:Ansatz_2}
\end{figure}

\begin{figure}[ht!]
\includegraphics[width=.4\textwidth]{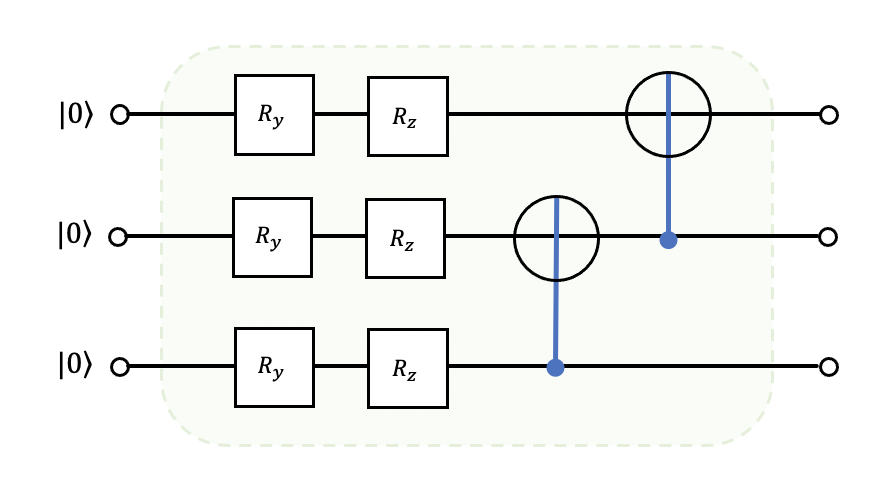}
\caption{$R_y$ and $R_z$ Ansatzes with $CNOT$ entanglement used in qCBR.}
\label{fig:Ansatz_3}
\end{figure}

Being  $\theta$ the vector of the parameters and  $R_{x}$  and  $R_{y}$  the unit gates of qubits used to create the Ansatz. To complement the experimentation scenario, it would be necessary to add the CNOT gate and the CRZ, which are the gates that help to achieve entanglement as seen in figure \eqref{fig:Ansatz_1}, \eqref{fig:Ansatz_2} and \eqref{fig:Ansatz_3}.


\begin{figure}[ht!]	
\includegraphics[width=.5\textwidth]{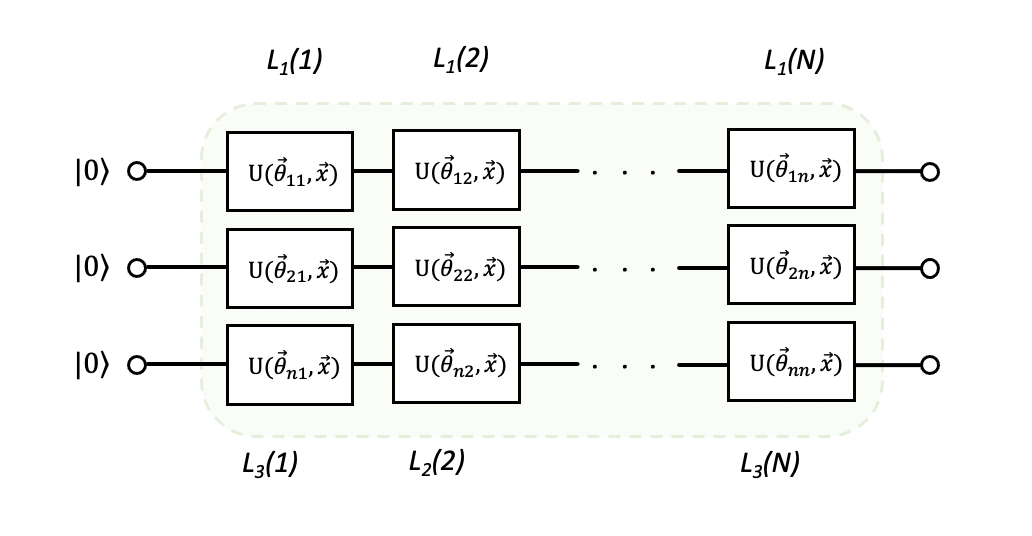}
\caption{Three-qubit quantum classifier circuit without entanglement.}
\label{fig:Ansatz_Comb}
\end{figure}

\begin{figure}[ht!]	
\includegraphics[width=.5\textwidth]{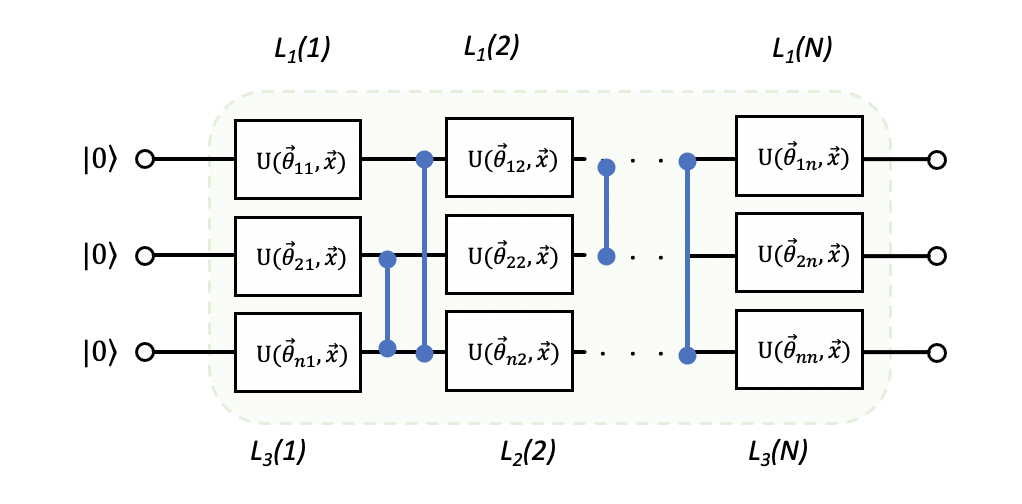}
\caption{Three-qubit quantum classifier circuit with entanglement by using $CZ$ or $CNOT$  gates.}
\label{fig:Ansatz_Comb_1}
\end{figure}

The variational quantum classifier structure (figure \eqref{fig:Ansatz_Comb} and \eqref{fig:Ansatz_Comb_1}) is based on layers of trainable circuit blocks  $ L \left( i \right) = \prod_{i,j}^{}U \left( i,j \right)$ and data coding, as shown in \eqref{Ansatz} for 8 dimensional or in \eqref{anstaz_1D} for 2 dimensional data size. Additionally, the entanglement can be achieved using the CRZ or CNOT gates.
The number of parameters to optimize the classifier is given by \eqref{num_parameter}.
\begin{equation}
\label{num_parameter}
    NumParam= ( nL\ast 2d \ast L) 
\end{equation}
In this case, with  $n$  the number of qubits,  $n = 3$ , $L$ the number of layers (blocks), in the experiment, it is a variable data and  $d$  which is the dimension of the problem. In other words, $d$ varies with the choice of Ansatz and whether or not entanglement is applied. In the case of the entanglements in figure \eqref{fig:Ansatz_2}, the $d$ would be summed 1 ($CRZ$ gate has one parameter), which equates to equation \eqref{num_paramenter_nD}.
\begin{equation}
\label{num_paramenter_nD}
NumParam= ( nL\ast 2 ( d+1 ) \ast L ) 
\end{equation}
\subsection{Fidelity cost function}\label{sec:App-Fidelity}
The similarity function follows the same strategy as the re-uploading and path; nevertheless, the Ansatz is different. It uses the definition of quantum fidelity associated with several qubits and maximizes said average fidelity between the test state and the final state corresponding to its class. Equation \eqref{Cost_Funtion} \cite{Adr20} defines the cost function used.
\begin{equation}
\label{Cost_Funtion}
\small
\begin{split}
    & CF \left( \overrightarrow{ \alpha },\overrightarrow{ \theta },\overrightarrow{ \omega } \right)  = \\ & = \frac{1}{2} \sum _{ \mu =1}^{M} \sum _{c=1}^{C}  \left(  \sum _{q=1}^{Q} \left(  \alpha _{c,q}F_{c,q} \left( \overrightarrow{ \theta },\overrightarrow{ \omega ,}\overrightarrow{x}_{ \mu } \right) -Y_{c} \left( \overrightarrow{x}_{ \mu } \right)  \right) ^{2} \right) 
\end{split}
\end{equation}
with
\begin{equation}
\label{fidelity_cost_function}
F_{c,q} \left( \overrightarrow{ \theta },\overrightarrow{ \omega ,}\overrightarrow{x}_{ \mu } \right) =\langle \psi _{c} \vert   \rho _{q} \left( \overrightarrow{ \theta },\overrightarrow{ \omega ,}\overrightarrow{x} \right)  \vert   \psi _{c} \rangle 
\end{equation}

Where  $\rho _{q}$  is the reduced density matrix of the qubit to be measured,  $M$  is the total number of training point,  $C$  is the total number of the classes,  $\overrightarrow{x}_{ \mu }$ are the training points and  $\overrightarrow{ \alpha }= \left(  \alpha _{1}, \ldots , \alpha _{C} \right)$  are introduced as class weights to be optimized together with  $\overrightarrow{ \theta } $,  $ \overrightarrow{ \omega ,}$  are the parameters and  $Q$  the numbers of the qubits. Counting on  $Y_{c} \left( \overrightarrow{x}_{ \mu } \right)$  as the fidelity vector for a perfect classification. This cost function \eqref{Cost_Funtion} is weighted and averaged over all the qubit that form this classifier.
In order to complete the hybrid system, it is used for the classical part, the following minimization methods above cited: L-BFGS-B \cite{ich95}, COBYLA \cite{The21} and SPSA \cite{Jam01}.

\section{The qCBR's details}\label{sec:App-CBR}
 The operation of the \textbf{retrieve} (prediction) block is given by a new case (schedule). In this experimentation, the schedule that best adapts to the latest case to be solved is recovered with the predict method, which is executed at a time  $O(log(MN))$. It worth saying that, due to the SWP descriptions, a possible schedule change, a stage of understanding or interpretation is necessary, since an adequate resolution of the new schedule cannot be carried out if it is not understood with some completeness. This stage of understanding is a simple decision algorithm with minimal intelligence.
 
 Once having the predicted solution, the synthesis block creates a new solution (proposed solution) by combining recovered solutions. To do this, the algorithm is divided into two main lines (figure \eqref{fig:Funct_qCBR}). A line that determines an acceptable degree of error (after a probabilistic study) that the predicted solution can be considered the proposed solution. The second branch is in charge of improving the expected solution towards a better-proposed solution. To do this, the \textit{Initial\_point} associated with the retrieved schedule is retrieved from the case memory, and the Variational Quantum Eigensolver is executed with very few shots, (\textit{k shots}). The idea here is to refine the new schedule's similarity with the recovered one. Operating the VQE with \textit{Initial\_point} provides the algorithm with parameter values through the initial point as a starting point for searching for the minimum eigenvalue (similarity between the two times) when the new time's solution point is believed to be close to a matter of the recovered schedule. This is how the \textbf{Re-use} block works. These operations have a complexity of $O(klog(N)+log(NM))$. Where  $N$  is the number of social workers, $M$ is the number of patients and $k$  is the number of shots.
 
 The algorithm's processes to review the proposed solution are seen below the \textbf{Re-use} block in figure \eqref{fig:Funct_qCBR}. It is essential to classify the best possible solution for the proposed prototype. The best possible solution is calculated with the VQE with the maximum resolution and depth (for the variational part). Once the solution is obtained, it is compared with the proposed solution and said solution with its characteristics is added to the new schedule before storing it (see figure \eqref{fig:qMem} and \eqref{fig:mem1}). The computational complexity of the Revise is determined by $O(log(N)+ICA))$. In this work, access to data (states) is determined by $O(log(MN))$ due to the characteristics of the inner products and superpositions.
 
 One of the most critical blocks in this work is to \textbf{Retain}. This block is the heart of the CBR because it is the classifier and because it is the block that allows us to conclude that it has been learned from the previous cases. Not all instances (schedules) are saved in this job, leading to the excessively slow classifier. Therefore, in this part of the algorithm, the best cases (timetables) that summarize all the essential information are retained.

\begin{figure}[!ht]
\centering
\includegraphics[width=0.5\textwidth]{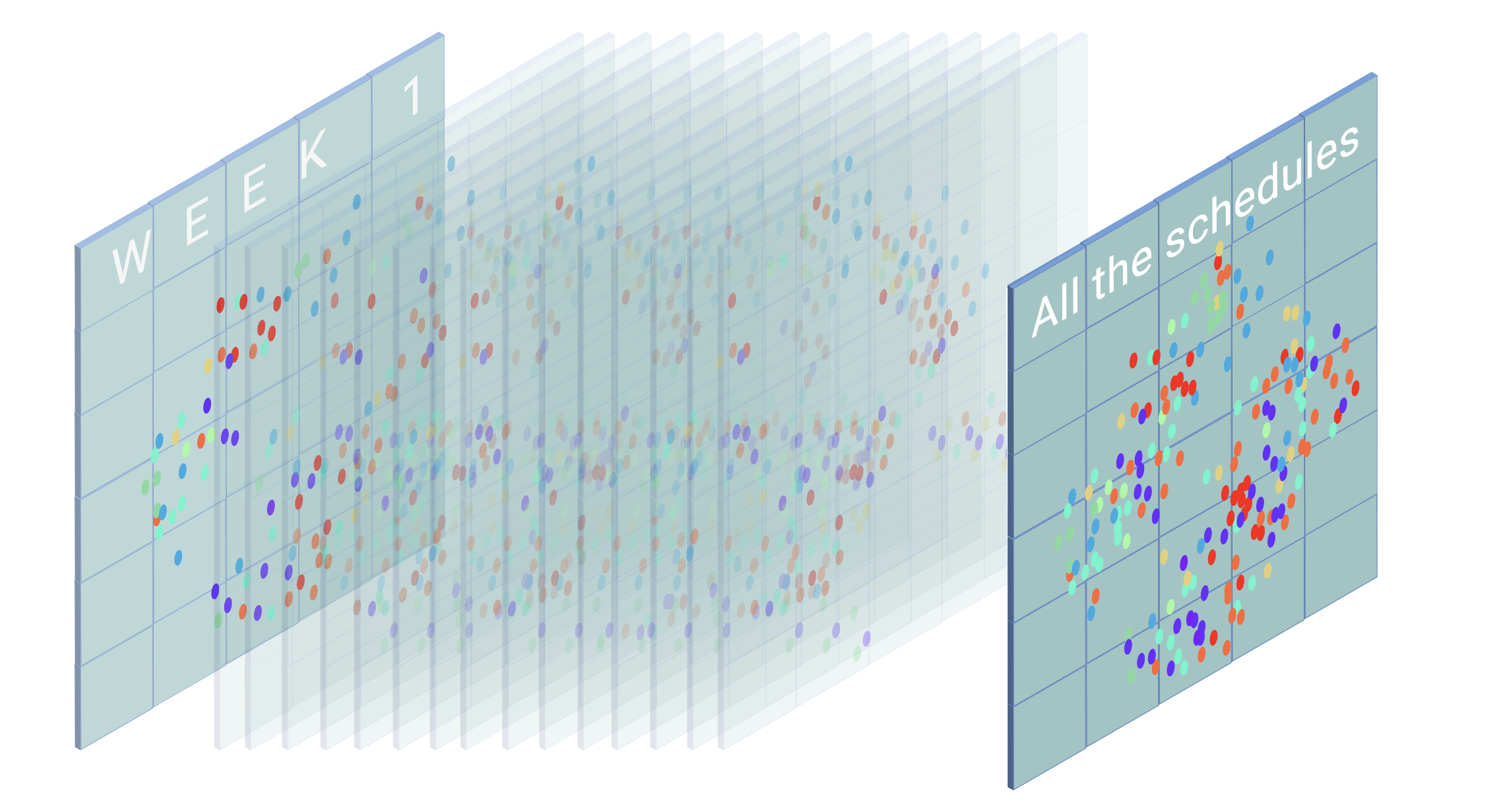}
\caption{Representation of the generation of the  $n$  weekly schedules of the SWP. The last plane, the one to the right of everything, represents all the SWP classes. The overlapping effect generated by the social workers' problem's characteristics and experimentation scenarios can be observed. This experimentation leads us to use the ICA technique to have a resulting dataset regardless of the schedules.}
\label{fig:Representation_SWP}
\end{figure}

\begin{figure}[!ht]
\centering
\includegraphics[width=0.5\textwidth]{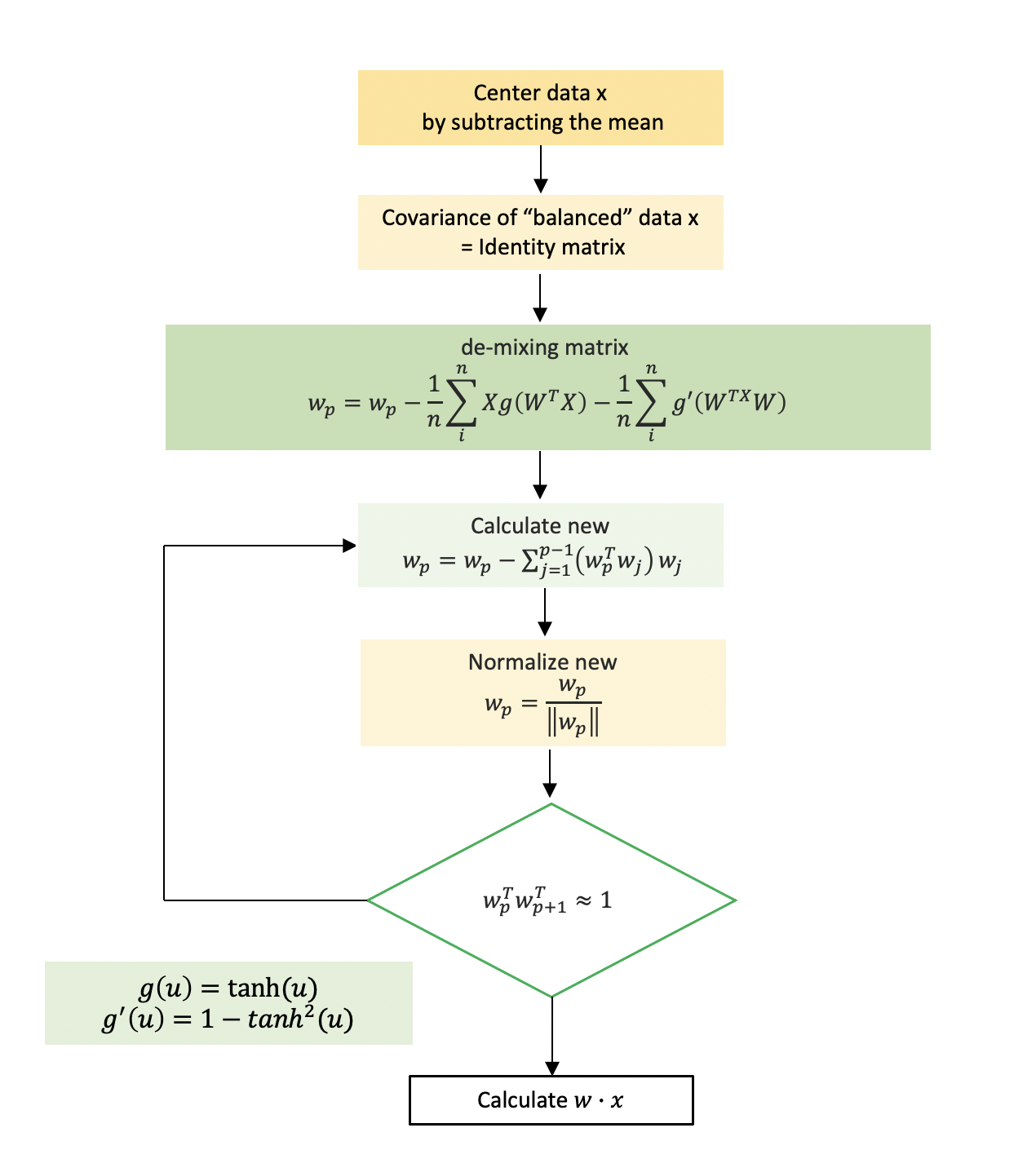}
\caption{Fundamental processes to apply ICA to SWP dataset. The first thing that is done is to centre the data x by subtracting the mean, balancing the data x removing its variance, and calculating the unmixing matrix of W. is calculated. Then the new value of w is calculated, and then w is normalized before checking if with the said value the algorithm converges or not. If it does not converge, a new w must be recalculated, and if it converges, calculate the scalar product of $\langle x,y \rangle$ to obtain the independent weekly schedules.}
\label{fig:ICA}
\end{figure}

The \textbf{Retain} process begins with the treatment of schedules, searching for the algorithm's best efficiency, which is a challenge to solve in this block.
In the case of SWP, the patient visits hours have a margin range of 30 min. Therefore, if one schedule starts at 9:00, the next could begin at 9:30, leading to a dataset with overlap between schedules if many schedules have similar time ranges spread over different days of the week. In the case of non-linearity of the data, an almost perfect classifier with an average accuracy more significant than $ 80 \% $ would be needed to be combined with a data processing system and a decision tree.

In this work, we contemplate both scenarios. First, get an excellent classifier and apply data processing techniques to help a poor classifier.
Using the standard classifier, ICA \cite{Hyv01_interscience} is applied to the original data to reduce the effects of the degree of overlap (figure \eqref{fig:Representation_SWP}) without losing the fundamental characteristics of the data. Figure \eqref{fig:ICA} summarizes the processes and operations applied to reduce the overlapping effect observed in the generation of SWP schedules. The complexity of this operation is noted as  $ O(ICA)$. The PCA is then used to reduce the data dimension from 8 to 2 and apply it to the designed variational classifier with the complexity equal to $O(PCA)$.
Once the best time is determined, we retain the knowledge acquired at the time of the case's resolution.
\newpage

\bibliographystyle{unsrturl}
\bibliography{main}

\end{document}